\documentclass{article}

\usepackage{graphicx}
\usepackage{comment}
\usepackage{amsmath,amssymb,amstext} %

\usepackage{gensymb}
\belowdisplayskip \abovedisplayskip

\newlength{\sectionReduceTop}
\newlength{\sectionReduceBot}
\newlength{\subsectionReduceTop}
\newlength{\subsectionReduceBot}
\newlength{\abstractReduceTop}
\newlength{\abstractReduceBot}
\newlength{\captionReduceTop}
\newlength{\captionReduceBot}
\newlength{\subsubsectionReduceTop}
\newlength{\subsubsectionReduceBot}

\newlength{\eqnReduceTop}
\newlength{\eqnReduceBot}

\newlength{\horSkip}
\newlength{\verSkip}

\newlength{\figureHeight}
\usepackage{relsize}
\usepackage[T1]{fontenc}
\usepackage[scaled=0.90]{inconsolata}
\usepackage[latin1]{inputenc}
\usepackage[english]{babel}

\usepackage{epsfig}
\usepackage{graphicx}
\usepackage{wrapfig}
\usepackage[belowskip=2pt,labelfont=bf,aboveskip=0pt,font=small,labelsep=period]{caption}
\usepackage{array}

\usepackage[belowskip=0pt,aboveskip=0pt,font=small]{subcaption}
\usepackage{longtable}
\usepackage{microtype}

\usepackage{textcomp}
\usepackage{stmaryrd}
\SetSymbolFont{stmry}{bold}{U}{stmry}{m}{n}
\usepackage{upgreek}
\usepackage{bm}
\usepackage{cases}
\usepackage{mathtools}
\usepackage{arydshln}
\usepackage{multirow}

\usepackage{color}
\usepackage[dvipsnames]{xcolor}
\definecolor{JalapenoRed}{RGB}{183,21,64}
\definecolor{Belize}{RGB}{41,128,185}
\definecolor{Amour}{RGB}{238,82,83}

\usepackage{cite}
\usepackage[pagebackref=true,breaklinks=true,colorlinks,bookmarks=false,allcolors=JalapenoRed]{hyperref}

\usepackage{algorithm}
\usepackage{algpseudocode}

\usepackage{multirow}
\usepackage{rotating}
\usepackage{booktabs}
\usepackage{etoolbox}
\newcounter{magicrownumbers}
\preto\tabular{\setcounter{magicrownumbers}{0}}
\newcommand\rownumber{\stepcounter{magicrownumbers}\arabic{magicrownumbers})\,}

\usepackage{enumitem}
\usepackage[olditem,oldenum]{paralist}

\usepackage{alltt}
\usepackage{listings}

\usepackage{url}
\usepackage{xspace}
\usepackage{comment}
\usepackage{afterpage}
\usepackage{pdfpages}
\usepackage{framed}
\usepackage{fancybox}
\usepackage{cuted}
\usepackage{caption}
\usepackage{hyperref}

\usepackage{quoting}
\quotingsetup{vskip=0pt}
\usepackage{epigraph}
\usepackage[normalem]{ulem}

\setdefaultleftmargin{.2cm}{.2cm}{}{}{}{}
\setlist{leftmargin=.2cm}

\usepackage[final]{corl_2020} %

\newtoggle{arxiv} %

\usepackage{mysymbols}
\iftoggle{arxiv}{}{\abovedisplayskip 3.0pt plus2pt minus2pt%
\belowdisplayskip \abovedisplayskip

\usepackage{letltxmacro}

\LetLtxMacro{\oldsection}{\section}
\renewcommand{\section}[1]{
    \vspace{-0.08in}
    \oldsection{#1}
    \vspace{-0.07in}
}

\LetLtxMacro{\oldsubsection}{\subsection}
\renewcommand{\subsection}[1]{
    \vspace{-0.06in}
    \oldsubsection{#1}
    \vspace{-0.05in}
}

\LetLtxMacro{\oldsubsubsection}{\subsubsection}
\renewcommand{\subsubsection}[1]{
    \oldsubsubsection{#1}
}

\setlength{\floatsep}{1ex plus 0ex minus 0.5ex}
\setlength{\textfloatsep}{3ex plus 0ex minus 0ex}
\setlength{\abovecaptionskip}{2ex plus 0ex minus 0ex}
\setlength{\belowcaptionskip}{0ex plus 0ex minus 0ex}}

\title{Auxiliary Tasks Speed Up Learning \\\pointnavfull}

\author{
    Joel Ye$^{1}$\thanks{Correspondence to \texttt{joel.ye@gatech.edu}}\quad
    Dhruv Batra$^{2,1}$\quad
    Erik Wijmans$^{1}$\thanks{EW and AD contributed equally.}\quad
    Abhishek Das$^{1\shortrightarrow2\dagger}$ \\
    $^1$Georgia Institute of Technology\quad
    $^2$
    Facebook AI Research\\
}

\begin{document}
\maketitle

\begin{abstract}

\pointnavfull is an embodied task that requires agents to navigate to a
specified point in an unseen environment.~\citet{wijmans_iclr20} showed that this
task is solvable in simulation but their method is computationally prohibitive --
requiring $2.5$ billion frames of experience and $180$ GPU-days. We develop
a method to significantly improve sample
efficiency in learning \pointnav using self-supervised auxiliary tasks
(\eg predicting the action taken between two egocentric observations,
predicting the distance between two observations from a trajectory, \etc). We find that naively combining multiple auxiliary tasks improves sample efficiency,
but only provides marginal gains beyond a point.
To overcome this, we use attention to combine representations from
individual auxiliary tasks. Our best agent is $5.5$x faster to match the
performance of the previous state-of-the-art, DD-PPO~\citep{wijmans_iclr20}, at
$40$M frames, and improves on DD-PPO's performance at $40$M frames by $0.16$
SPL. Our code is publicly available at
\href{https://github.com/joel99/habitat-pointnav-aux}{{\tt github.com/joel99/habitat-pointnav-aux}}.


\end{abstract}

\section{Introduction}
\label{sec:intro}

Consider a robot tasked with navigating from the bedroom
to the kitchen solely from first-person egocentric vision. To do so,
it must be able to reason about 1) notions of free space (that
doors can be walked through, but not walls), 2) keep regions already
visited in memory (so as not to run around in circles), 3) common sense of how
houses and objects are typically laid out (that kitchens typically are not inside
bedrooms), \etc. To learn these skills, the agent needs a good environment representation.

The current state-of-the-art method for training a class of such robots
(embodied agents) in simulation
is Decentralized Distributed PPO (DD-PPO)~\citep{wijmans_iclr20}.
Specifically,~\citet{wijmans_iclr20} train an agent to autonomously navigate to
a point-goal in an unseen environment nearly perfectly ($99.9$\% success rate).
However, this comes at a prohibitive computational cost -- requiring $2.5$
billion frames of experience; $80+$ years of experience accrued
over \emph{half-a-year} of GPU time, 64 GPUs for 3 days!

While \citep{wijmans_iclr20} serves as an excellent `existence proof' of the
learnability of \pointnav, we believe it should not take $2.5$ billion frames of
experience and nearly 6 months of GPU time to learn to navigate from point A to B.
\iftoggle{arxiv}{
In mathematics and theoretical computer science, an existence proof is often
the first crack in the wall, frequently followed soon thereafter by improvements
to the underlying techniques -- a non-constructive proof replaced by constructive
proof, an improved algorithm, shaving off factors in bounds -- overcoming barriers till the problem is well-understood.
}{
An existence proof is often
the first crack in the wall, enabling subsequent improvements -- a non-constructive proof replaced by constructive
proof, an improved algorithm, shaving off factors in bounds -- until the problem is well-understood.
}
\emph{That} is our goal -- to improve sample and time efficiency in learning \pointnav using self-supervised auxiliary tasks.

\iftoggle{arxiv}{
These tasks (\eg predicting the action taken between two egocentric
observations, predicting the distance between two observations from a
trajectory, predicting future observations in a trajectory from current
observation) are `self-supervised' --~\ie make use of information already
available to the agent -- and
`auxiliary' --~\ie requiring core competencies independent of
any particular downstream task (such as \pointnavfull).
}{}

\begin{figure}[t]
    \captionsetup[subfigure]{font=scriptsize,labelfont=scriptsize}
    \centering
    \begin{subfigure}[c]{.42\textwidth}
        \centering
        \includegraphics[width=\linewidth]{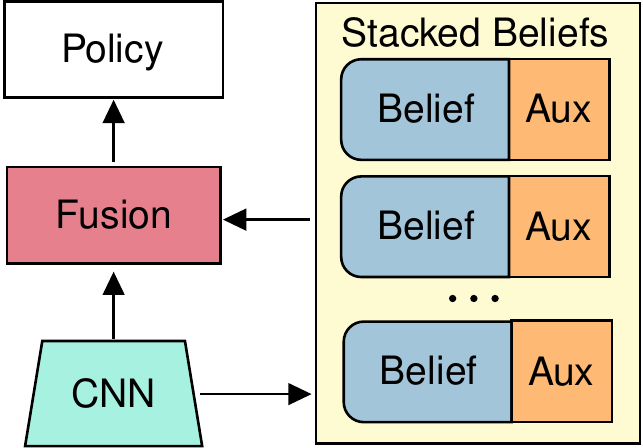}
        \vspace{0.9cm}
        \caption{}\label{fig:1b}
    \end{subfigure}%
    \hspace{0.1cm}
    \begin{subfigure}[c]{.52\textwidth}
        \centering
        \includegraphics[width=\linewidth]{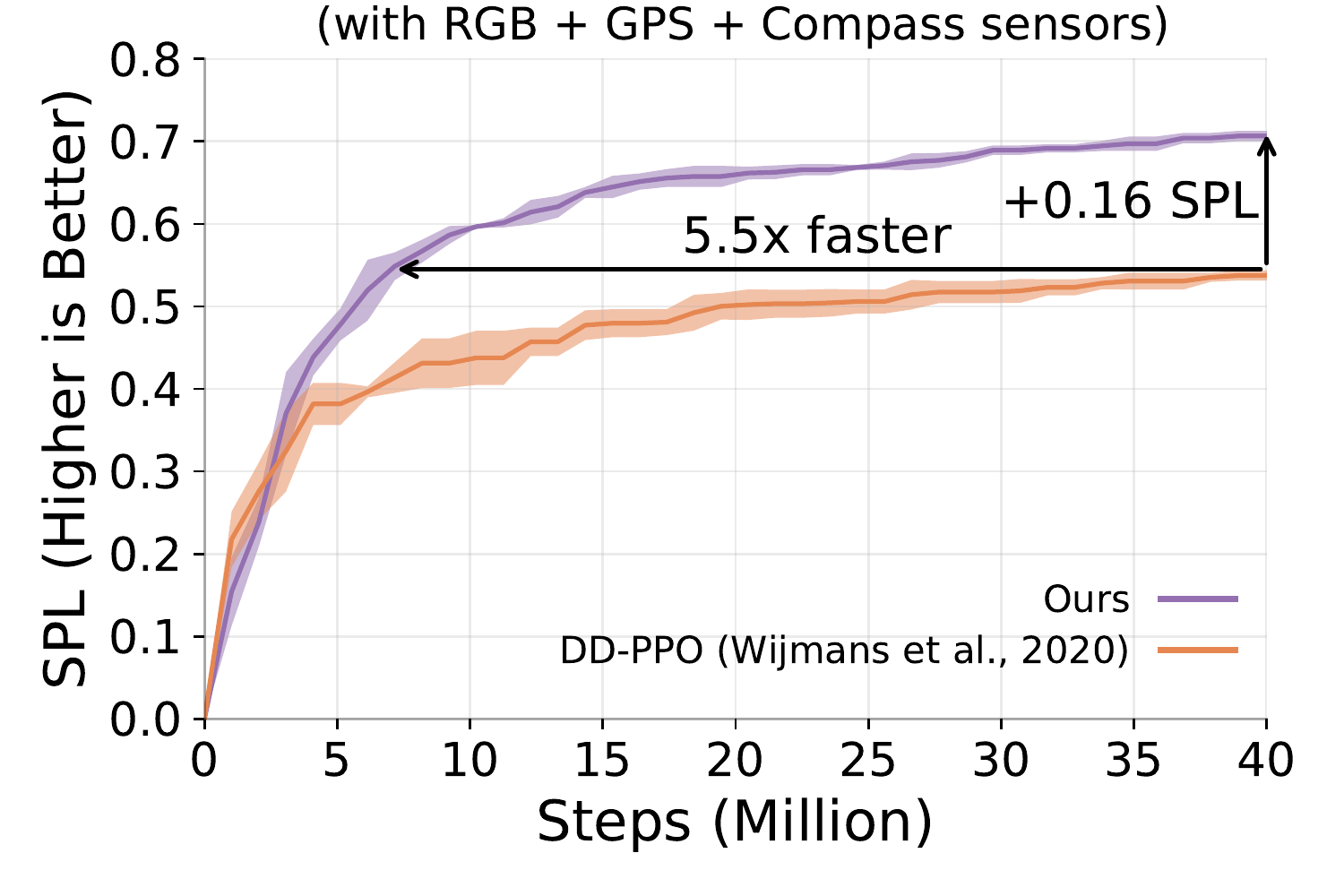}
        \caption{}\label{fig:1c}
    \end{subfigure}
    \caption{(a) We use learning signals from multiple self-supervised
        auxiliary tasks on a recurrent architecture (detailed in
        \secref{sec:aux-tasks-rollout}) to speed up learning
        \pointnav. (b) Our best agent achieves the same performance as the
        DD-PPO~\citep{wijmans_iclr20} baseline $5.5\times$ faster
        and improves on the baseline's performance at $40$M frames by $0.16$ SPL.}
    \label{fig:teaser}
\end{figure}

In the process of improving sample efficiency, we address several
important questions over prior work in auxiliary self-supervised learning,
from both the supervised~\citep{wang_iccv15,pathak_cvpr16,noroozi_eccv16,oord_arxiv18,ma_iclr19,bachman_neurips19,tian_arxiv19,he_arxiv19,misra_arxiv19,chen_arxiv20,chen_xinlei_arxiv20}
and reinforcement learning paradigms~\citep{jaderberg_arxiv16,guo_arxiv18,pathak_icml17,gregor_neurips19,anand_neurips19,lin_nips19,mirowski_arxiv16, hjelm_iclr2019,gordon_iccv19}.
First, auxiliary tasks are typically benchmarked in visually simple
simulated environments (\eg DeepMind Lab~\citep{beattie_arxiv16}, Atari).
Do these improvements transfer to realistic environments?
Second, it is unclear how these auxiliary objectives interact with each other --
can multiple such tasks be combined? Do they lead to positive transfer when combined,
or is there interference? Finally, what is the `right' way to combine them -- can
they be naively combined by summing the losses, or does combining them necessitate
sophisticated weighting mechanisms?

\newpage
Concretely, our contributions are the following:
\begin{compactitem}[--]
    \item We \emph{significantly} improve sample- and time-efficiency on \pointnavfull over DD-PPO~\citep{wijmans_iclr20}.
    \item We study three self-supervised auxiliary tasks --
        action-conditional contrastive predictive coding (\cpc)~\citep{guo_arxiv18},
        inverse dynamics\iftoggle{arxiv}{(predicting the action taken between two egocentric
        observations)~\citep{pathak_icml17}}{~\citep{pathak_icml17}}, and temporal distance estimation --
        and show that each \iftoggle{arxiv}{of these objectives}{} improves sample
        efficiency over the baseline agent from~\citep{wijmans_iclr20}.
        With a fixed computation budget (of $40$M steps of experience), our best single
        auxiliary task \cpcat4 improves performance from $0.55$ to $0.66$ SPL ($+22$\%).
        With a fixed performance level ($0.55$ SPL), \cpcat16 achieves a $2.1\times$
        reduction in no. of steps required (from $40$M to $19$M).
    \item Next, we show that the naive combination (\ie direct addition of losses)
        of multiple auxiliary tasks can further improve sample efficiency over
        single tasks. The best combination achieves $.70$ SPL ($+27$\%) by $40$M
        steps and achieves $0.55$ SPL in $12$M steps,
        a $3.3\times$ speedup.
    \item Finally, we observe that naive summation of losses has
        diminishing returns on sample efficiency as we further increase the number of auxiliary tasks.
        We propose a novel
        attention mechanism to fuse state representations that overcomes these
        negative effects.
        Putting it all together, our final model obtains $0.55$ SPL in
        $7$M steps of experience, a $5.5\times$ speedup over the baseline from~\citep{wijmans_iclr20}.
\end{compactitem}
\section{Related Work}

Our work relates to prior work in auxiliary objectives
for learning representations in reinforcement learning, methods for combining
multiple such objectives, and other approaches to PointNav.

\xhdr{Auxiliary Tasks in Reinforcement Learning.}
Auxiliary tasks provide additional complementary objectives to
improve sample efficiency and/or performance on the primary task.
Supervised auxiliary tasks expose privileged information to the agent
(such as depth~\citep{mirowski_arxiv16,embodiedqa,eqa_modular}).
Self-supervised auxiliary tasks, such as next-step visual feature prediction~\citep{pathak_icml17},
predictive modeling~\citep{guo_arxiv18,gregor_neurips19}, or spatio-temporal mutual information
maximization~\citep{hjelm_iclr2019,anand_neurips19} derive supervision from the
agent's own experience. In contrast to prior work,
which focus on simpler and non-photorealistic
environments~\citep{mirowski_arxiv16,guo_arxiv18,gregor_neurips19,anand_neurips19,lin_nips19},
we focus on visually complex, photorealistic environments from
the Gibson 3D scans~\citep{xia_cvpr18}.%

Closely related to our work is that of~\citet{gordon_iccv19} who show
that auxiliary tasks can be leveraged to improve transfer to new tasks and new
simulation environments (\ie synthetic to photorealistic). In contrast, we
focus on improving sample efficiency when learning a task \emph{from scratch},
proposing that the most performant representations should arise by
virtue of end-to-end learning.

\xhdr{Combining Multiple Auxiliary Tasks.} Combining multiple auxiliary tasks raises the challenges that
1) they have varying affinities with a
given primary task, and 2) these affinities can change during the training
process as the agent improves.
When studying knowledge transfer between multiple tasks, most prior work comes
from multi-task learning. There, task affinity has often been taken as a constant,
where influence is normalized by task uncertainty~\citep{kendall_cvpr18}, or as
a prior~\citep{evgeniou_jmlr2005}.
In contrast, we propose a formulation that \emph{learns the appropriate influence}
of each auxiliary task during training.

\citet{lin_nips19} uses gradient similarity to
adaptively weight auxiliary losses. However, this approach is limited
to training time, whereas our approach also enables auxiliary task weighting during \emph{evaluation}.~\eg,
long-horizon predictive modeling (\myquote{what room is my goal in?}) may be useful overall, but inappropriate when
turning a tight corner. While such an ability could be implicit in a
loss-driven approach,
\iftoggle{arxiv}{we propose an explicit approach
using adaptive weights over multiple `experts' (representations learnt
via separate auxiliary tasks). An explicit weight}
{an explicit weight}
distribution sidesteps the need for mathematical approximations as in~\citep{lin_nips19} and
allows for inference-time visualization of task influence.

\xhdr{PointGoal Navigation.}
PointGoal Navigation (detailed in \secref{sec:task_sim_agent}) has progressed remarkably, with several entries to the 2019 Habitat PointNav Challenge exceeding $0.70$ SPL in only $10$M observations. One leading method is
Active Neural Mapping~\citep{chaplot_iclr20},
which uses neural environment maps and hierarchical planning modules. \citet{sax_arxiv18} transferred visual
features from Taskonomy~\citep{zamir_cvpr18}, showing no single representation was ideal for multiple embodied tasks and concluding diverse representation sets are best for unknown downstream tasks. \citet{shen_iccv19}
presented a vision-conditioned fusion of visual representations that outperformed naive concatenation. Our work operates in the same regime, using dot-product attention~\citep{vaswani_nips17} to guide
fusion.
As transferred visual representations have been promising, we briefly compare with~\citep{sax_arxiv18} in~\ref{sec:results_prior}. However, our work seeks to improve
``from-scratch'' training of \textsc{PointNav} agents, significantly simplifying
the training pipeline. Thus our contributions are orthogonal to these approaches~\eg, our approach can improve the local planner in~\citep{chaplot_iclr20} (see~\secref{sec:local_appendix}).
As for~\citep{shen_iccv19}, we note their fusion technique is inherently sample
inefficient. Each of their policies must be trained individually before fusion can
be used, which means samples required scale linearly with the number of tasks fused.
Further, their approach uses many large ResNet-50 encoders, while our approach has
a footprint smaller by over 100x FLOPs. Our main comparison is with~\citep{wijmans_iclr20},
where from-scratch representations effectively solved \pointnav.

\section{Task, Simulation, Agent\iftoggle{arxiv}{, and Training}{}}
\label{sec:task_sim_agent}

\xhdr{\pointnavfull.}
In \pointnav~\citep{anderson_arxiv18}, an agent is initialized in an \emph{unseen}
environment and tasked with navigating to a goal location without a map.
The goal location is specified with coordinates relative to initial location
(\eg `go to (5, 20)' where units describe distance relative to start in meters). The agent is equipped with an \texttt{RGB} camera (providing egocentric \texttt{RGB}
observations) and a GPS+Compass sensor (providing position and orientation
relative to the start location).
The agent has access to 4 standard actions: $\{\texttt{move forward}\,(0.25\text{m}),
\,\texttt{turn left}\,(10\degree),\,\texttt{turn right}\,(10\degree),\,\texttt{stop}\}$.

\xhdr{Metrics.}
We evaluate the agent on two metrics -- 1) Success: whether or not the agent
correctly predicted \texttt{stop} within $0.2$m of the goal, and 2) Success
weighted by inverse Path Length (SPL)~\citep{anderson_arxiv18}: which weights
success by how efficiently the agent navigated to the goal relative to the shortest path.

\xhdr{Simulation.} We simulate our agent on the AI Habitat platform~\citep{savva_iccv19}, which has been shown to have good Sim2Real transfer~\citep{kadian2019making}.
Following the 2019 Habitat Challenge~\citep{savva_iccv19},
we train and evaluate performance on the higher quality reconstructions from the Gibson dataset~\citep{xia_cvpr18}, \ie $72$ houses for training and $14$ houses for validation. We test our approach's ability to generalize in~\secref{sec:mp3d}.

\begin{figure}[t]
    \centering
    \includegraphics[width=12cm]{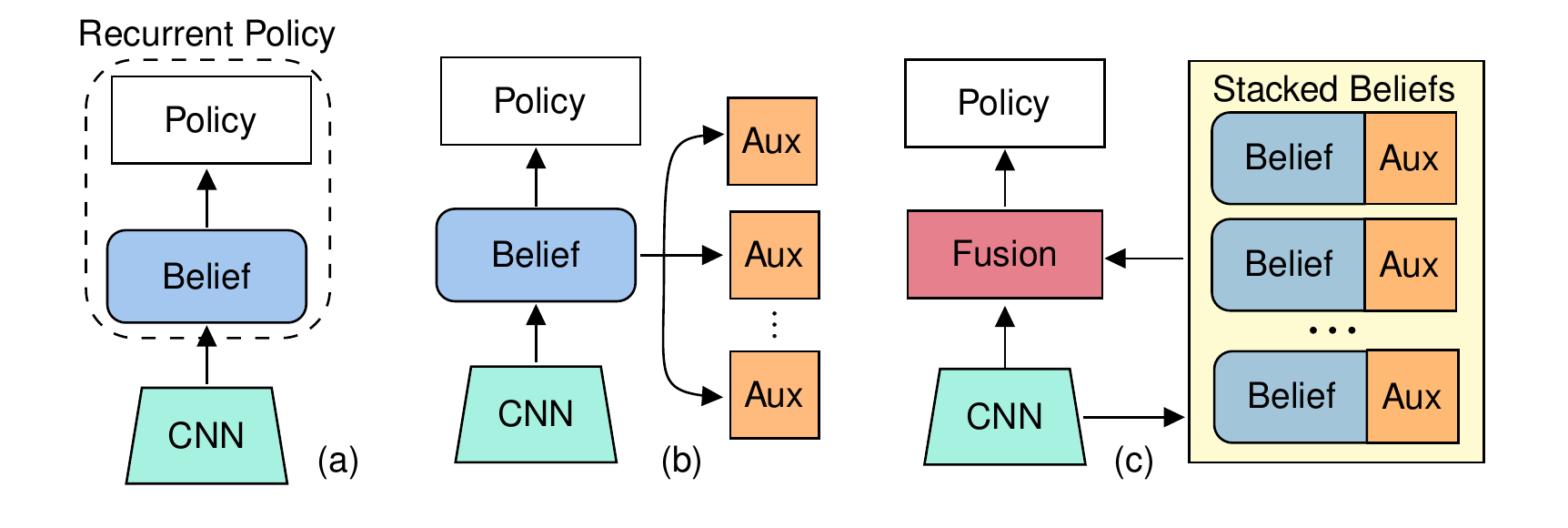}
    \caption{
        (a) Baseline architecture from DD-PPO~\citep{wijmans_iclr20} with the
        recurrent policy conceptually separated into a recurrent `belief' module
        and a feed-forward policy head.
        (b) Single-belief, where multiple auxiliary tasks utilize the same
        shared belief module output.
        (c) Fused-beliefs, where each auxiliary task is paired with its own
        separate belief module, and the outputs of the belief modules are fused
        and fed as input to the policy head.}
    \label{fig:architecture}
\end{figure}

\xhdr{Agent.}  We divide our agent into three separate components: a convolutional neural network (CNN) encoder that produces an embedding of the visual observation (\texttt{RGB}), a `belief' module that integrates multiple observations to produce an actionable summary representation, and a policy head that determines the agent's action given the belief module output. Note that prior work commonly refers to this architecture as consisting of two parts -- a CNN encoder and a recurrent policy.
We divide this recurrent policy into a recurrent belief module and a feedforward policy head as shown in~\figref{fig:architecture}a. We denote this split as our auxiliary tasks operate on belief module output, as shown in~\figref{fig:architecture}b.

Our modifications to the baseline architecture are intentionally minimal, isolating the impact of auxiliary tasks. We use ResNet18~\citep{he_cvpr16} as modified for on-policy RL by~\citet{wijmans_iclr20} for our visual encoder.

The belief module is a single layer GRU~\citep{cho_emnlp14}. Its output $h_t$ passes to the policy head, a
fully-connected layer, to yield a softmax distribution over the action space and
a value estimate.

\iftoggle{arxiv}{
\xhdr{Training.}  We train our agent via Proximal Policy Optimization
PPO)~\citep{schulman_arxiv17} with Generalized Advantage Estimation
(GAE)~\citep{schulman_iclr16}. We use 4 rollout workers with rollout length $T=128$,
and 4 epochs of PPO with 2 mini-batches per epoch.
We set discount factor to $\gamma=0.99$ and GAE factor $\tau=0.95$.
We use the Adam optimizer~\citep{kingma_iclr15} with a learning rate of $2.5\times10^{-4}$ and $\epsilon=0.1$.s
We follow the reward structure in~\citep{savva_iccv19}. For goal $g$,
when the agent is in state $s_t$ and executes action $a_t$ (transitioning to $s_{t+1}$),
\begin{equation}
    r_t(s_t, a_t) = \begin{cases}
    2.5 \cdot \text{Success} & \text{if } a_t = \texttt{stop} \\
    \texttt{GeoDist}(s_{t}, g) -
    \texttt{GeoDist}(s_{t+1}, g) - \lambda & \text{otherwise}
    \end{cases}
\label{eq:pnreward}
\end{equation}
where \texttt{GeoDist} is the geodesic distance and
$\lambda ($=$0.01)$ is a slack penalty.
}{}

\section{Self-Supervised Auxiliary Tasks from Experience}
\label{sec:aux-tasks-rollout}

\begin{figure}[t]
    \centering
    \includegraphics[width=0.95\textwidth]{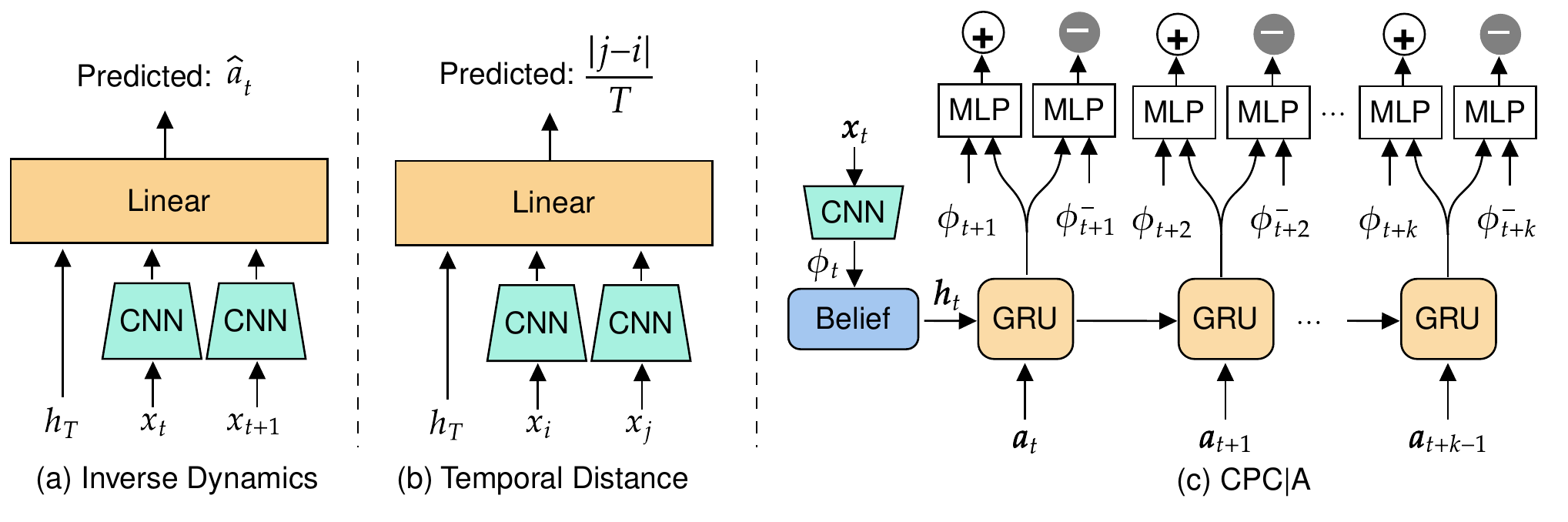}
    \caption{We study three auxiliary modules.
    a) Inverse dynamics: decoding action taken from successive visual embeddings
    $\phi_t$ and $\phi_{t+1}$ and the final belief state $h_T$.
    b) Temporal distance: decoding the timestep difference between two observation
    embeddings from final belief state $h_T$.
    c) \cpc: contrasting future observation embeddings ($\phi_{t+1}, ...,  \phi_{t+k}$)
        at every timestep from other observation embeddings using a secondary GRU.}
    \label{fig:id-td-cpc}
\end{figure}

We introduce a set of auxiliary modules, one for each auxiliary task.
\iftoggle{arxiv}{While the policy head is trained to maximize rewards for the primary task -- \pointnav (Eq.~\ref{eq:pnreward})
-- these auxiliary modules provide additional learning signals
and operate on the observations, outputs of the belief module, and actions.
}{The auxiliary modules operate on observations, outputs of the belief modules, and actions.}
Specifically, the agent receives observation $x_t$,
extracts its CNN representation $\phi_t$,
which is fed to the belief module to compute $h_t$,
used to sample an action $a_t$ from the policy.
Auxiliary tasks use a subset of $\{(x_1,\phi_1,h_1,a_1)\dots(x_T, \phi_T, h_T, a_T)\}$.
Our choice of auxiliary tasks is motivated by providing the agent the ability to learn environment dynamics
(which actions separate two observations?, how would the environment look if I moved forward and turned right?, \etc).
We specifically only consider self-supervised tasks (that do not need additional supervision)
\iftoggle{arxiv}{}{in~\figref{fig:id-td-cpc}}, to keep the method generally
applicable in simulation and the real world. This disallows tasks like depth prediction,
which is known to make \pointnav much easier to learn~\citep{chaplot_iclr20,mirowski_arxiv16}.
\iftoggle{arxiv}{
We next describe the tasks in detail.
}{
We describe the tasks in detail in~\secref{sec:aux_detailed}. In experiments, we consider \cpc with
$k=1,2,4,8,16$ (\cpcat1, \cpcat2 \etc). The entire family is denoted as \allcpc.
}

\iftoggle{arxiv}{
\xhdr{Inverse Dynamics (ID).}
As shown in~\figref{fig:id-td-cpc}a, given two successive
observations ($x_t$ and $x_{t+1}$) and the belief module hidden state at the
end of the trajectory ($h_T$), the ID task is to predict the action taken at time
$t$, $a_t$. We include the belief module hidden state to encourage representation
of trajectory actions in it.

\xhdr{Temporal Distance (TD).}
As shown in~\figref{fig:id-td-cpc}b, given two observations from a trajectory ($x_i$ and $x_j$) and
the belief module hidden state at the end of the trajectory ($h_T$),
the TD task is to predict $\frac{|j-i|}{T}$.
This is similar in spirit to progress estimation in~\citep{ma_iclr19} and
reachability in~\citep{savinov_iclr18}.
However, rather than Euclidean or geodesic distance, we ask the agent to predict the
(normalized) number of steps elapsed between two visual observations.
This requires the agent to recall if a location is revisited or similarly viewed,
designed to promote understanding of spatio-temporal relations of trajectory viewpoints.

\xhdr{Action-Conditional Contrastive Predictive Coding (\cpc)}.
As shown in~\figref{fig:id-td-cpc}c,
given the belief module hidden state $h_t$, a second GRU is unrolled for $k$ timesteps using future actions $\{a_{t+i}\}_{i=0}^{k-1}$ as input.
The output of the second GRU at time $t+i$ is used to distinguish different visual representations.
We concatenate the second GRU's output at $t+i$ with a) the ground-truth visual representation at $t+i$, $\phi_{t+i+1}$, or b) a ``negative'' visual feature $\phi_{t+i+1}^-$ sampled from other timesteps and trajectories. Then, "contrasting" the different representations can be framed as a classification task, classifying inputs with the ground-truth visual representation as $1$, and negatives as $0$.

This encourages $h_t$ to build long-horizon representations of the environment.
In our experiments, we consider the instances of \cpc with
$k=1,2,4,8,16$ (denoted \cpcat1, \cpcat2 \etc), which leads to a family of 5 \emph{similar} tasks,
each being optimized for representations that make predictions at different
timescales. We denote the entire family as \allcpc.

}{}

During training, we optimize the parameters of the visual encoder, belief module, and policy head, altogether $\theta_m$,
as well as the auxiliary module parameters, $\theta_a$, to jointly minimize the
auxiliary loss and the primary \pointnav objective, $L_\text{RL}$, with $\beta_{\text{Aux}}$ as a hyperparameter balancing the losses:
\begin{align}
    L_{\text{total}}(\theta) & = L_\text{RL}(\theta_m) + \beta_{\text{Aux}} L_\text{Aux}(\theta_m,\theta_a)
\end{align}
We set $\beta_{\text{Aux}}$ such that the two losses have roughly equal magnitudes at
initialization.

\subsection{Leveraging Multiple Auxiliary Tasks}
\label{sec:multiauxnaive}

If individual auxiliary tasks help, a natural question to ask is whether their improvements are additive.
A simple approach is to apply different tasks to the single belief module (see~\figref{fig:architecture}b), adding all the individual loss terms
with the same loss scales as in the individual task setup.
For $n_{\text{Aux}}$ such auxiliary tasks, we denote individual auxiliary
task-related parameters as $\theta_a^1\dots \theta_a^{n_{\text{Aux}}}$. The
new loss is given by:
\begin{equation}
     L(\theta_m;\theta_a^1\dots\theta_a^{n_{\text{Aux}}}) = L_{\text{RL}}(\theta_m) + \sum_{i=1}^{n_{\text{Aux}}} \beta^i_{\text{Aux}} L_{\text{Aux}}(\theta_m;\theta_a^i)
\label{eq:sumaux}
\end{equation}

\iftoggle{arxiv}{\iftoggle{arxiv}{}{
\begin{equation}
     L(\theta_m;\theta_a^1\dots\theta_a^{n_{\text{Aux}}}) = L_{\text{RL}}(\theta_m) + \sum_{i=1}^{n_{\text{Aux}}} \beta^i_{\text{Aux}} L_{\text{Aux}}(\theta_m;\theta_a^i)
\label{eq:sumaux-supplement}
\end{equation}
}

\iftoggle{arxiv}{\xhdr{Weighted \cpcat16.}}{}
Following 
\iftoggle{arxiv}{ Eq.~\ref{eq:sumaux} }{Eq.~\ref{eq:sumaux-supplement} (reproduced for reference)}
with \allcpc as auxiliary
tasks leads to a setting where 1-next step prediction gets counted $5$ times
in the overall loss function (once each across $k=1, 2, 4, 8, 16$), 2-step
predictions get counted $4$ times (once each across $k=2, 4, 8, 16$), and so on.
Since all the \allcpc tasks are structurally similar, we can reduce computation by
emulating this total loss in a single auxiliary task. We do this via a single
`weighted \cpcat16', where 1-step prediction is scaled by $5$,
2-step prediction is scaled by $4$, and so on.
}{}

\subsection{Attention over multiple auxiliary tasks}
\label{sec:multiauxattn}

\begin{wraptable}{R}{0pt}
\vspace{-15pt}
     \centering
     \scriptsize
     \begin{tabular}{p{2.5cm} p{4.5cm}}
     Fusion Method & Description \\
     \midrule
     \raggedright Fixed~\citep{pathak_icml17} &
     Full weight fixed to a single belief module.
     \\
     \raggedright Average &
     Equal weighting on all belief modules.\\
     \raggedright Softmax Gating \citep{shen_iccv19} &
     Visually conditioned softmax weighting on belief modules. $\mathbf{w} = \text{softmax}(f(\mathbf{\phi}))$.
     \\
     \raggedright Scaled Dot-Product Attention (Attn)\citep{vaswani_nips17}  &
     Weights computed as $\text{score}(h_i, k)=\frac{h_i^Tk}{\sqrt{n_\text{Aux}}}$
     with beliefs $h_i$ and $\mathbf{k} = g_{\text{key}}(\mathbf{\phi})$.

     \\
     \bottomrule
    \end{tabular}
    \caption{All fusion methods are a weighted sum of beliefs. See~\secref{sec:fusion_details} for further details.}
    \label{table:fusion}
\vspace{-15pt}
\end{wraptable}

\iftoggle{arxiv}{As we investigate in~\secref{sec:experiments}, we find that using
the setup described in~\secref{sec:multiauxnaive} to combine auxiliary
tasks does improve performance over individual tasks, but the gains are marginal beyond a point.
}{As we investigate in~\secref{sec:experiments}, using the method in~\secref{sec:multiauxnaive}
to combine auxiliary tasks does improve performance over single tasks, but
gains quickly diminish.}
We hypothesize that this is because when multiple auxiliary tasks operate on the same
belief module (\figref{fig:architecture}b), their objectives compete.
Additional objectives can thus hinder learning.
To better leverage multiple auxiliary tasks, we propose a novel
architecture with a shared CNN, separate recurrent belief modules for each auxiliary
task, and a `fusion' module to combine belief module outputs into an input for the policy head, depicted in~\figref{fig:architecture}c. With separate
belief modules, auxiliary tasks can optimize their respective beliefs
for orthogonal objectives without interference, and the fusion module can
extract policy-relevant representations.
\iftoggle{arxiv}{
    \begin{figure}[t]
        \centering
        \includegraphics[width=12cm]{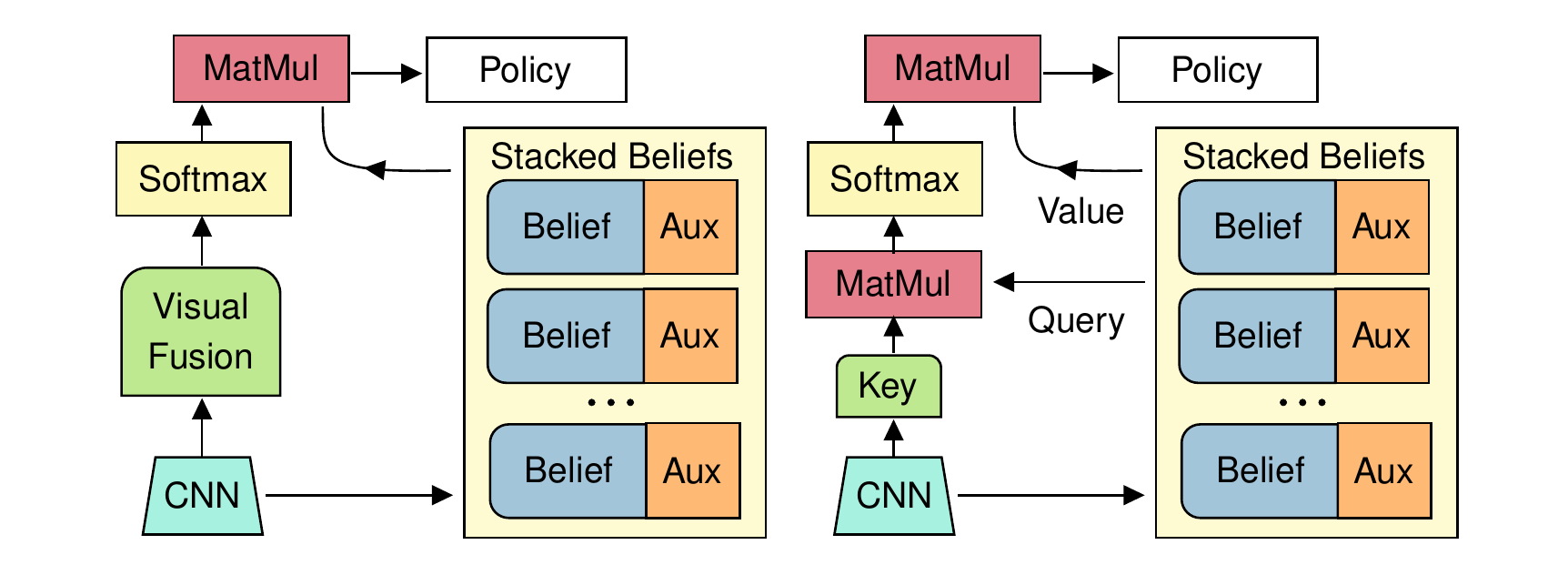}
        \caption{We use two learned weightings to combine different belief
            representations.
            Left: Softmax gating predicts weights conditioned on the visual representation.
            Right: Scaled dot-product attention computes scores by taking dot
            product of belief module outputs with a key vector computed from the
            visual representation, normalized by $\sqrt{n_{\text{Aux}}}$.}
        \label{fig:fusion}
    \end{figure}

We experiment with several fusion methods (\tableref{table:fusion}).
Softmax gating and scaled dot-product attention are shown in
\figref{fig:fusion}, with details in~\secref{sec:fusion_details}.
We additionally consider a variant of scaled dot-product attention
with an entropy penalty (denoted `\textsc{+E}'). Given attention distribution
$w_{\text{attn}}\vcentcolon= (p_1,\dots,p_{n_{\text{Aux}}})$, we calculate entropy
as $\sum_{i=1}^{n_{\text{Aux}}} -p_i\log p_i$. Entropy encourages the agent to
use multiple belief modules. An agent that quickly learns to use a single module may prevent the other modules from learning about the task (from reduced gradients).
}{We experiment with several fusion methods (\tableref{table:fusion}, details in~\secref{sec:fusion_details}).
We further apply an entropy penalty on the attention distribution (denoted `\textsc{+E}'),
to encourage the use of multiple modules (details in~\secref{sec:fusion_details}).}
\iftoggle{arxiv}{
\input{sections/experiments_v2}
}{
\section{Experiments and Results}
\label{sec:experiments}

We aim to answer the following questions:

\begin{enumerate}[leftmargin=0cm,itemindent=.5cm,labelwidth=\itemindent,labelsep=0cm,align=left]
    \item Do auxiliary tasks help on \pointnav in photorealistic environments?
    \item Does combining auxiliary tasks help over individual tasks?
    \item What is the best way to fuse representations from multiple auxiliary tasks?
\end{enumerate}

We refer to observations as `frames' of simulation throughout, as done in prior work. Each variant
(\secref{sec:aux-tasks-rollout}) is trained for $40$M frames as this corresponds to
$1$ GPU-week and with $4$ random seeds. We report the highest average
validation (averaged over three validation runs) SPL achieved
by $40$M frames. Note that validation is performed on held-out scenes and reward is not available during evaluation.  Analyzing success shows similar trends as SPL, so we reserve those results for~\secref{sec:experiments-appendix}.
To analyze sample efficiency, we compare the area under the learning curves (AuC) over $40$M frames,
with measurements every $1$M frames. Models with higher AuC learn
\pointnav faster, an important skill for more challenging tasks, where
slow learning might be intractable. When computing AuC, we first normalize the
x-axis (no. of frames) to $[0.0, 1.0]$, which normalizes AuC to the same range.
In tables, we bold one variant over others with overlapping confidence intervals
if it has better performance across validation episodes (paired t-test, $p < 0.05$).

\begin{figure}[t!]
    \centering
    \begin{subfigure}{.43\textwidth}
    \resizebox{\textwidth}{!}{
        \begin{tabular}{@{}lccc@{}}
            & AuC & Best \\
            \toprule
            \rownumber Baseline &
            $0.422 $\scriptsize{$\pm 0.043$} & $0.545 $\scriptsize{$\pm 0.010$}
            \\[0.01in]
            \rownumber \cpcat$1$ &
 $0.480 $\scriptsize{$\pm 0.032$} & $0.628 $\scriptsize{$\pm 0.011$}
\\
\rownumber \cpcat$2$ &
 $0.487 $\scriptsize{$\pm 0.035$} & $0.641 $\scriptsize{$\pm 0.008$}
\\
\rownumber \cpcat$4$ &
 $0.512 $\scriptsize{$\pm 0.029$} & $0.658 $\scriptsize{$\pm 0.014$}
\\
\rownumber \cpcat$8$ &
 $0.517 $\scriptsize{$\pm 0.027$} & \boldsymbol{$0.658 $}\scriptsize{\boldsymbol{$\pm 0.012$}}
\\
\rownumber \cpcat$16$ &
 $0.514 $\scriptsize{$\pm 0.029$} & \boldsymbol{$0.663 $}\scriptsize{\boldsymbol{$\pm 0.010$}}
\\
\rownumber ID&
 $0.458 $\scriptsize{$\pm 0.036$} & $0.588 $\scriptsize{$\pm 0.011$}
\\
\rownumber TD &
 $0.441 $\scriptsize{$\pm 0.044$} & $0.564 $\scriptsize{$\pm 0.017$}
\\
            \bottomrule
         \end{tabular}
    }
    \caption{}\label{tab:results-simple}
    \end{subfigure}%
    \begin{subfigure}{.57\textwidth}
      \centering
      \includegraphics[width=\linewidth]{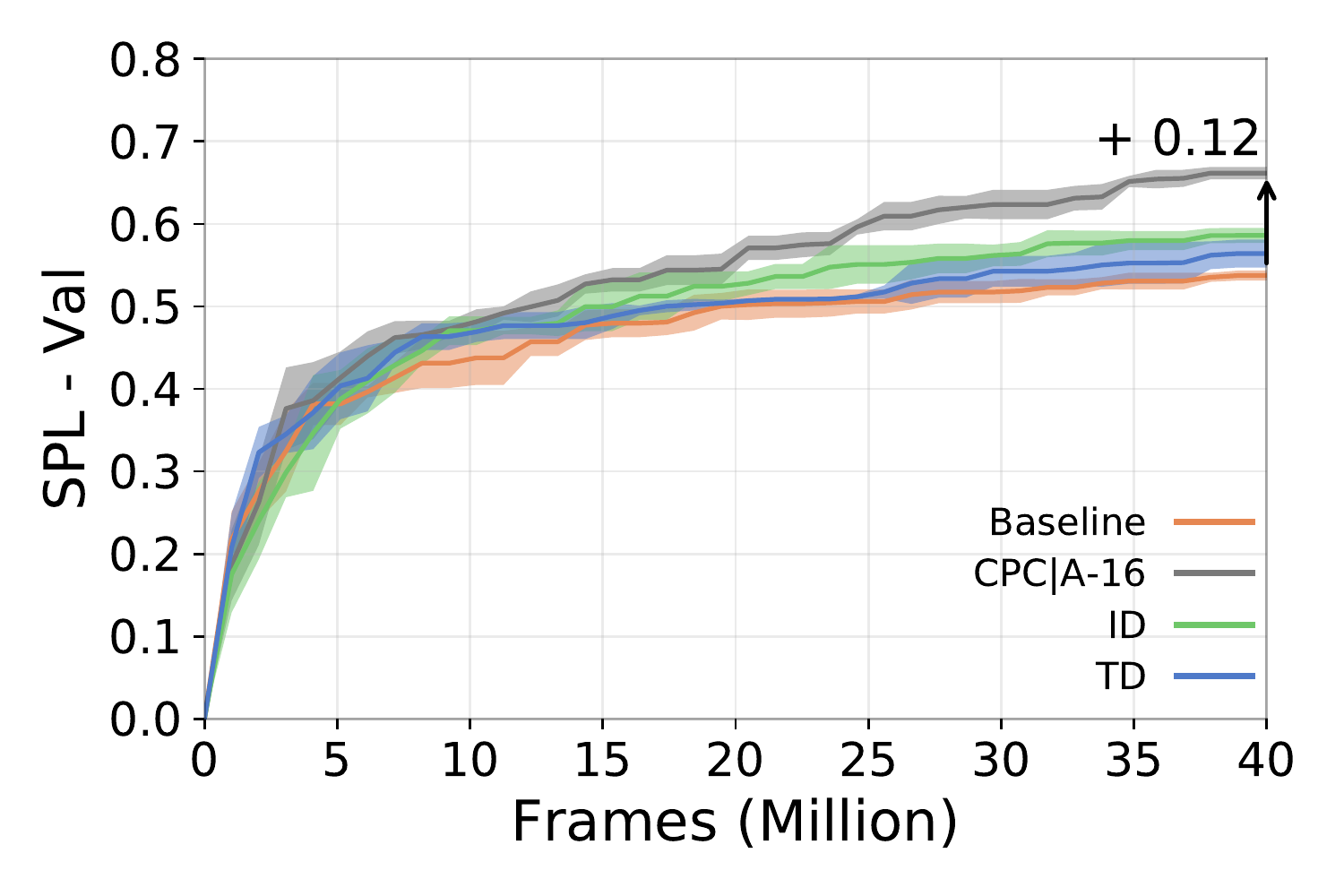}
      \caption{}\label{fig:simple_plots}
    \end{subfigure}
    \caption{(a) Auxiliary tasks accelerate learning of \pointnav. Long-range \cpc tasks provide more gain, \eg \cpcat16 provides +0.12 SPL (+22\%) by 40M frames. (b) Auxiliary tasks overtake the baseline by 5M frames.
    }
\end{figure}

All variants with single auxiliary tasks get higher SPL than the baseline, as shown in~\tableref{tab:results-simple}. \cpc excels at
longer ranges, \ie rows 4-6 indicate they provide at
least $+0.11$ ($+20\%$) SPL at 40M frames, and $+0.09$ ($+21\%$) SPL AuC. The slight edge longer ranges have over \cpc-\{1,2\} is consistent with intuitions in~\citep{gregor_neurips19}.
\iftoggle{arxiv}{
With similar metrics in $k=4,8,16$, we
use $k=4$ as the best single task in subsequent comparisons.
}{We subsequently
use $k=16$ as reference for the best single task.}

All variants, including the baseline, have a ramp-up after which metrics begin to level (\eg $\sim0.5$ SPL), as seen in~\figref{fig:simple_plots}. Intuitively, this inflection point
represents when an easier subset of episodes is learned, and subsequent episodes
provide diminishing returns. The dropoff is softer for variants with auxiliary tasks,
indicating where better representations are benefiting primary task learning.
Relatedly, we observe a subtle  early performance loss for modified variants in~\figref{fig:simple_plots}. They overtake the baseline as it levels
($4$M frames). Initial
learning involves the agent's first successes, requiring understanding the existence of the goal point and its large reward. Intuitively, these navigational auxiliary
tasks are distracting from the rare initial success rewards and it does not pay
off for the agent to learn better representations before the agent has latched
on to the success reward.

The efficacy of these tasks ($+10\%$ training time,
$+20\%$ SPL) already motivates use in future baselines.

\subsection{Adding multiple auxiliary tasks improves over single tasks}
\begin{figure}[t!]
    \centering
    \begin{subfigure}{.57\textwidth}
    \resizebox{\textwidth}{!}{
        \begin{tabular}{@{}lccc@{}}
            & AuC & Best \\
            \toprule
            \rownumber Baseline & 
 $0.422 $\scriptsize{$\pm 0.043$} & $0.545 $\scriptsize{$\pm 0.010$}
\\[0.01in]
\cdashline{1-4}
\rownumber \cpcat$16$ (Best Single) & 
 $0.514 $\scriptsize{$\pm 0.029$} & $0.663 $\scriptsize{$\pm 0.010$}
\\[0.01in]
\cdashline{1-4}
\rownumber \allcpc: Add & 
 $0.523 $\scriptsize{$\pm 0.026$} & \boldsymbol{$0.687 $}\scriptsize{\boldsymbol{$\pm 0.010$}}
\\
\rownumber \allcpc+ID+TD: Add & 
 $0.532 $\scriptsize{$\pm 0.028$} & \boldsymbol{$0.696 $}\scriptsize{\boldsymbol{$\pm 0.013$}}
\\
            \bottomrule
         \end{tabular}
    }
    \caption{}\label{tab:results-homo}
    \end{subfigure}%
    \begin{subfigure}{.43\textwidth}
      \centering
      \includegraphics[width=\linewidth]{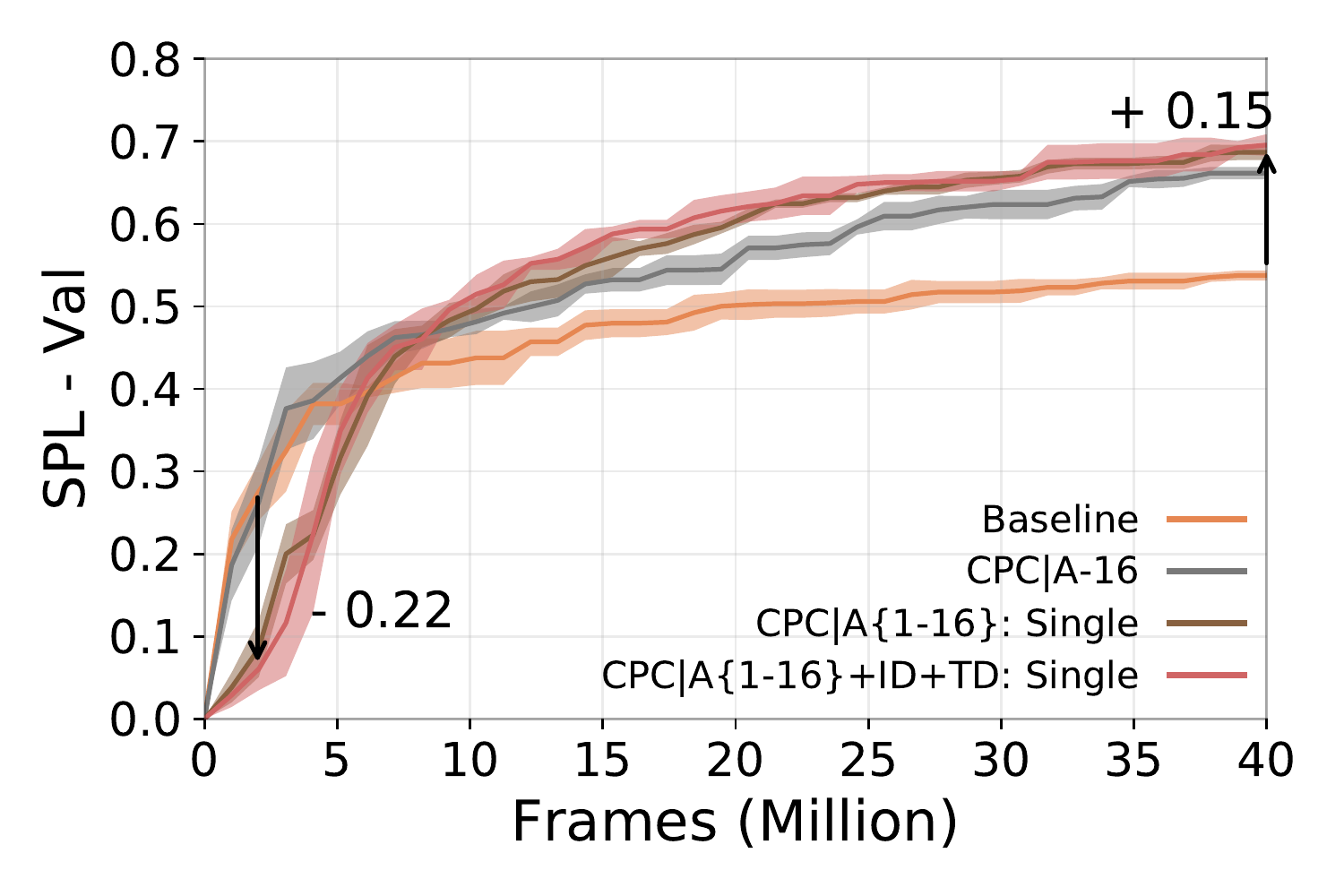}
      \caption{}\label{fig:five_plots}
    \end{subfigure}
    \caption{(a) \allcpc provides +0.02 SPL over \cpcat16, and adding ID+TD yields only \textbf{+0.01 SPL more}. (b) Using multiple auxiliary tasks improves on a single task by 10M frames at cost to initial ramp-up.
    }
\end{figure}

Given the improvements from individual auxiliary tasks,
we next assess if these improvements are complementary
by naively combining these tasks, as given by Eq.~\eqref{eq:sumaux}. These experiments are shown in~\tableref{tab:results-homo}. We experiment with combining similar auxiliary tasks, using all \cpc variants (\allcpc), and also separately add ID and TD to diversify tasks used.
As these variants add task losses (as in~\figref{fig:architecture}b), we refer to them as `Add' in figures.

A sharper early performance loss is clear in~\figref{fig:five_plots}
\allcpc: Add and \allcpc+ID+TD: Add, compared to the baseline. Multiple task variants only
surpass the baseline at ${\sim}5$M frames, likely due to interference with
the primary \pointnav task. Using all CPC tasks (row 3) does bump SPL by +0.02, though further adding ID and TD (row 4) yields a minor marginal gain. This is surprising, as ID and TD
should be providing learning signals distinct from \cpc tasks.
We hypothesize that distinct learning signals interfere with each other
when using a single belief module.

\subsection{Attention over belief modules outperforms naive summation}
\label{sec:attention_belief}
\begin{figure}[t!]
    \centering
    \begin{subfigure}{.57\textwidth}
    \resizebox{\textwidth}{!}{
        \begin{tabular}{@{}lccc@{}}
            & AuC & Best \\
            \toprule
            \rownumber Baseline &                           
    
            $0.422 $\scriptsize{$\pm 0.043$} & $0.545 $\scriptsize{$\pm 0.010$}
            \\[0.01in]
            \rownumber \cpcat$16$ (Best Single) &
            $0.514 $\scriptsize{$\pm 0.029$} & $0.663 $\scriptsize{$\pm 0.010$}
            \\[0.01in]
            \cdashline{1-4}
            \rownumber \allcpc+ID+TD: Add &
            $0.532 $\scriptsize{$\pm 0.028$} & $0.696 $\scriptsize{$\pm 0.013$}
            \\[0.01in]
            \cdashline{1-4}
            \rownumber \allcpc+ID+TD: Average & 
 $0.539 $\scriptsize{$\pm 0.022$} & $0.696 $\scriptsize{$\pm 0.008$}
\\
\rownumber \allcpc+ID+TD: Soft\cite{shen_iccv19} & 
 $0.578 $\scriptsize{$\pm 0.020$} & $0.696 $\scriptsize{$\pm 0.008$}
\\
\rownumber \allcpc+ID+TD: Attn & 
 $0.565 $\scriptsize{$\pm 0.027$} & $0.698 $\scriptsize{$\pm 0.015$}
\\
\rownumber \allcpc+ID+TD: Attn+E & 
 $0.594 $\scriptsize{$\pm 0.019$} & \boldsymbol{$0.707 $}\scriptsize{\boldsymbol{$\pm 0.006$}}
\\
            \bottomrule
        \end{tabular}
    }
    \caption{}\label{tab:results-diverse}
    \end{subfigure}%
    \begin{subfigure}{.43\textwidth}
      \centering
      \includegraphics[width=\linewidth]{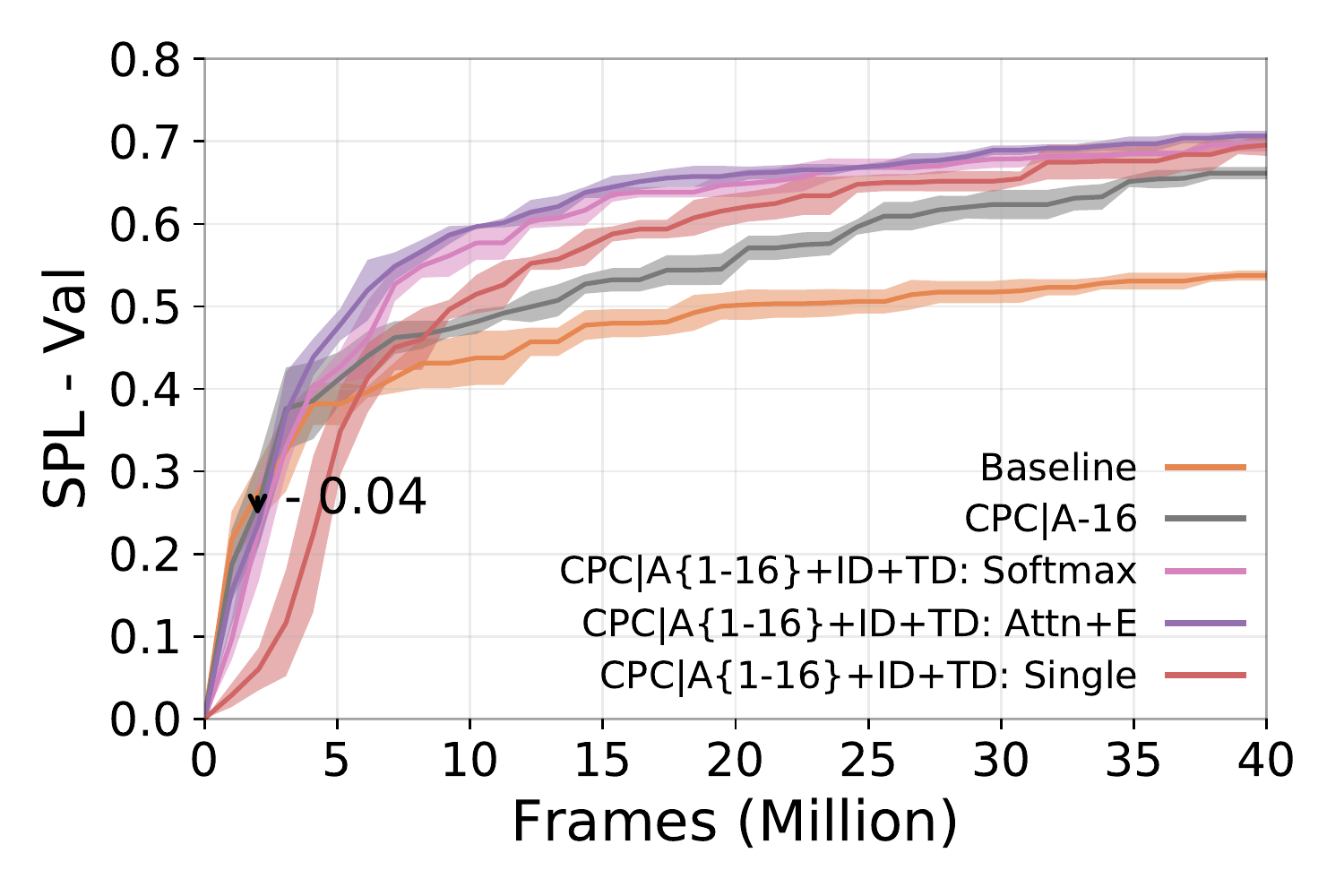}
      \caption{}\label{fig:seven_plots}
    \end{subfigure}
    \caption{(a) Learned fusion (row 7) provides a +0.01 SPL over a single module (row 3). 
    (b) Learning fusion of separate modules in \allcpc+ID+TD: Attn+E reduces ramp-up cost from \allcpc+ID+TD: Add.}
\end{figure}

To minimize hypothesized task conflict,
we next describe experiments with fusing representations from multiple belief modules
(as described in Sec.~\ref{sec:multiauxattn}). Results are shown in~\tableref{tab:results-diverse}. Averaging module features (row 4) leads to very similar
performance as a single module (row 3), both in AuC and in final performance.
This is expected, as the two variants are similar from the policy head's perspective.
However, learned fusion yields significant gains in AuC over a single module, \ie softmax fusion (row 5)
and attentive fusion (row 6) achieve $+0.04$ SPL over single module (row 3).

Though these models have similar SPL by 40M frames, speeding up initial
learning may be critical for getting off the ground in harder tasks.
We propose learned fusion thus enables the use of multiple different signals
while mitigating slow initial learning. In fact, we see precisely this
in~\figref{fig:seven_plots} - both fusion methods match the initial pace of \cpcat16 (a single task), improving over \allcpc+ID+TD: Add.

In our experiments, we found that the variants using attention quickly
(in $<0.5$M frames) start attending to just one belief module,
preventing the other belief modules from influencing the policy.
This collapsed attention implies the attention variant's improvements at
$40$M frames are primarily due to an improved visual representation, as
unattended belief modules still backpropagate gradients to the CNN. In that case,
the belief modules are relying on the shared improved visual representation,
rather than forming distinct beliefs as intended. We thus rectify the attention collapse with an entropy penalty on the attention
distribution (see \secref{sec:training_details} for details). With the penalty, the agent consistently attends
to all its belief modules, resulting in row 7, the Attn+E variant.
With this modification, the Attn+E variant edges out the softmax's (row 5) AuC and best performance.

\subsection{Comparison with pre-trained weights}
\label{sec:results_prior}

\begin{wraptable}{r}{0pt}

\begin{tabular}{@{}lccc@{}}
    & AuC & Best \\
        \toprule
        \rownumber CPC|A-1 & $0.480 $\scriptsize{$\pm 0.032$} & $0.628 $\scriptsize{$\pm 0.011$}
\\[0.01in]
        \rownumber Depth (Fine-tuned) & $0.451 $\scriptsize{$\pm 0.045$} & $0.616 $\scriptsize{$\pm 0.033$}
        \\
        \rownumber Depth (Frozen) & $0.454 $\scriptsize{$\pm 0.037$}
        & $0.612 $\scriptsize{$\pm 0.012$}
        \\
        \bottomrule
    \end{tabular}
\caption{With an improved visual encoder and without access to a bitmap of past
    locations~\cite{sax_arxiv18}, an agent with one auxiliary task performs
    similarly to an agent using Taskonomy depth weights, with and without fine-tuning.}
\label{tab:results-prior}

\vspace{-10pt}
\end{wraptable}

We also compare with using Taskonomy~\citep{zamir_cvpr18} weights for the visual
encoder, specifically representations used for depth predictions, known to
transfer well to \pointnav~\citep{sax_arxiv18}. We report results with and
without finetuning the projection layer and final ResNet block
in \tableref{tab:results-prior}.

Contrary to~\citep{sax_arxiv18}, we find that neither variant outperforms our baseline.
This has two primary causes. First, our baseline is stronger. \citet{sax_arxiv18} used a simple 3-layer CNN~\citep{mnih2015human} for their from scratch baseline, as provided in the Habitat baselines repository~\citep{savva_iccv19}. We use ResNet18, improving on~\citep{savva_iccv19}'s results for \pointnav \texttt{RGB} by 0.1 SPL at 40M.
Second, our transfer results are lower than that of~\citep{sax_arxiv18} due to three differences in training procedure and agent design.
\begin{inparaenum}[1)]
\item ~\citet{sax_arxiv18} use an off-policy version of PPO that trades compute for sample efficiency,
\item do not learn the stop action, which makes the task easier, and
\item provide the agent with a bitmap of previously visited locations instead of a GRU.
\end{inparaenum}
This finding strengthens the case for learning representations from scratch rather than transferring from static visual tasks.

}
\section{Model Analysis}
\label{sec:analysis}

\iftoggle{arxiv}{
\subsection{Ablative Analysis}
\label{sec:ablative_analysis}

Given the impressive gains of learned fusion, we conduct additional experiments to decompose its improvements. We use the five \allcpc tasks to focus on the effects of attention over similar tasks. Specifically, we address
the following:
\vspace{-0.01in}
\begin{compactenum}
\item We verify improvements when attention is largely fixed on one belief module (a symptom of insufficient entropy) to be due to improvements in visual representation.
    To show this, we artificially fix the agent's attention such that it always attends
    to \cpcat1 -- ``Fixed Attn'' -- even though gradients from all auxiliary tasks in the \allcpc
    family backpropagate to the visual encoder.
\item Do separate belief modules help when auxiliary tasks are similar?
    We investigate the effects of attention even when our tasks are all \cpc variants.
\item Do similar auxiliary tasks offer distinct gains?
    We apply the same auxiliary task, \cpcat16, to 5 separate belief modules. We denote this \cpcat16$\times 5$ and compare to \cpcat16.
\end{compactenum}
\vspace{-0.01in}

Separately, we introduce a ``Weighted CPC'' task to assess the value of differentiating the CPC tasks.
\iftoggle{arxiv}{}{\iftoggle{arxiv}{}{
\begin{equation}
     L(\theta_m;\theta_a^1\dots\theta_a^{n_{\text{Aux}}}) = L_{\text{RL}}(\theta_m) + \sum_{i=1}^{n_{\text{Aux}}} \beta^i_{\text{Aux}} L_{\text{Aux}}(\theta_m;\theta_a^i)
\label{eq:sumaux-supplement}
\end{equation}
}

\iftoggle{arxiv}{\xhdr{Weighted \cpcat16.}}{}
Following 
\iftoggle{arxiv}{ Eq.~\ref{eq:sumaux} }{Eq.~\ref{eq:sumaux-supplement} (reproduced for reference)}
with \allcpc as auxiliary
tasks leads to a setting where 1-next step prediction gets counted $5$ times
in the overall loss function (once each across $k=1, 2, 4, 8, 16$), 2-step
predictions get counted $4$ times (once each across $k=2, 4, 8, 16$), and so on.
Since all the \allcpc tasks are structurally similar, we can reduce computation by
emulating this total loss in a single auxiliary task. We do this via a single
`weighted \cpcat16', where 1-step prediction is scaled by $5$,
2-step prediction is scaled by $4$, and so on.
}

\begin{table*}[t]
    \centering
    \resizebox{.975\textwidth}{!}{
        \begin{tabular}{@{}lccccc@{}}
            & \multicolumn{2}{c}{Area under Curve} & & \multicolumn{2}{c}{Best} \\
            \cline{2-3} \cline{5-6}
            & Success $(\mathbf{\uparrow})$ & SPL $(\mathbf{\uparrow})$ & & Success $(\mathbf{\uparrow})$ & SPL $(\mathbf{\uparrow})$ \\[0.05in]
            \toprule
            \rownumber Baseline & 
 $0.583 $\scriptsize{$\pm 0.061$} & $0.422 $\scriptsize{$\pm 0.043$}
& &
 $0.714 $\scriptsize{$\pm 0.011$} & $0.545 $\scriptsize{$\pm 0.010$}
\\
\rownumber \cpcat$16$ (Best Single) & 
 $0.677 $\scriptsize{$\pm 0.039$} & $0.514 $\scriptsize{$\pm 0.029$}
& &
 $0.819 $\scriptsize{$\pm 0.019$} & $0.663 $\scriptsize{$\pm 0.010$}
\\
\rownumber Weighted \cpcat16 & 
 $0.674 $\scriptsize{$\pm 0.030$} & $0.527 $\scriptsize{$\pm 0.026$}
& &
 $0.839 $\scriptsize{$\pm 0.015$} & $0.679 $\scriptsize{$\pm 0.009$}
\\
\rownumber \allcpc: Add & 
 $0.665 $\scriptsize{$\pm 0.028$} & $0.523 $\scriptsize{$\pm 0.026$}
& &
 $0.831 $\scriptsize{$\pm 0.007$} & $0.687 $\scriptsize{$\pm 0.010$}
\\
\rownumber \allcpc+ID+TD: Attn+E & 
 $0.756 $\scriptsize{$\pm 0.019$} & $0.594 $\scriptsize{$\pm 0.019$}
& &
 \boldsymbol{$0.854 $}\scriptsize{\boldsymbol{$\pm 0.008$}} & \boldsymbol{$0.707 $}\scriptsize{\boldsymbol{$\pm 0.006$}}
            \\[0.01in]
            \cdashline{1-6}
            
            \rownumber \allcpc: Attn & 
 $0.707 $\scriptsize{$\pm 0.025$} & $0.557 $\scriptsize{$\pm 0.022$}
& &
 \boldsymbol{$0.847 $}\scriptsize{\boldsymbol{$\pm 0.002$}} & $0.695 $\scriptsize{$\pm 0.010$}
\\
\rownumber \allcpc: Attn+E & 
 $0.712 $\scriptsize{$\pm 0.032$} & $0.560 $\scriptsize{$\pm 0.030$}
& &
 \boldsymbol{$0.843 $}\scriptsize{\boldsymbol{$\pm 0.017$}} & $0.692 $\scriptsize{$\pm 0.010$}
\\
\rownumber \allcpc: Fixed Attn & 
 $0.691 $\scriptsize{$\pm 0.029$} & $0.543 $\scriptsize{$\pm 0.027$}
& &
 \boldsymbol{$0.845 $}\scriptsize{\boldsymbol{$\pm 0.011$}} & \boldsymbol{$0.698 $}\scriptsize{\boldsymbol{$\pm 0.011$}}
\\
\rownumber \cpcat16$\times 5$: Attn & 
 $0.694 $\scriptsize{$\pm 0.029$} & $0.549 $\scriptsize{$\pm 0.023$}
& &
 $0.832 $\scriptsize{$\pm 0.006$} & $0.683 $\scriptsize{$\pm 0.011$}

            \\
            \bottomrule
        \end{tabular}
    }
    \caption{Performance of attentive fusion and ablations on \cpc family. \allcpc: Attn (rows 6, 7) increases AuC, as expected. All variants continue to converge to similar metrics at 40M observations.}
    \label{table:results-cpc-ablations}
\end{table*}

Our results are summarized in~\tableref{table:results-cpc-ablations}, and
we list main findings below.

\iftoggle{arxiv}{}{
Weighted \cpc (row 3) does manage to outperform our best \cpc task (row 2), capturing the intuition that predicting further timesteps are more valuable for building environmental dynamics. Nonetheless weighting does not reach the performance of separate task modules (row 4).
}

\xhdr{Improvements are partly due to better visual representations.}
We find that benefits of multiple auxiliary tasks are partly
due to improvement in visual representations
(that is, \tableref{table:results-cpc-ablations}, row 8 and 6 have similar best results).
This mirrors the findings of~\citet{sax_arxiv18}, who use visual encoders pretrained on mid-level vision tasks (\eg 3D curvature prediction) to improve sample efficiency. Notably, however, though using attention enables slightly improved AuC, indicating attention may enable adaptive learning of some form.

Also, \allcpc: Attn+E (row 7) again improves on \allcpc: Add in Success AuC by almost $7\%$. This reaffirms \xhdr{separate belief modules help}, even if auxiliary tasks are from the same family. Adding entropy appears unhelpful in this setting where the auxiliary tasks are highly similar. %

Finally, \xhdr{similar tasks are slightly better than identical tasks}.
We find that a variant using correlated but distinct tasks performs slightly better than applying the same auxiliary task to all belief modules (\allcpc: Attn~\vs~\cpcat16$\times 5$: Attn, rows 6 and 9 SPL). The slight distinction in signal provided does yield a better policy. %

\subsection{Examining usage of belief modules}
}{
Ablative analysis is done in~\secref{sec:ablative_analysis}. Here, we examine how the agent uses its belief modules.
}
Though different tasks provide quantitative differences, we would like to determine whether they induce characteristic ``beliefs'' in their modules, \ie different functional roles.
We study a run of our best variant (\allcpc+ID+TD: Attn+E) trained to $40$M frames.

To quantify each module's contribution to performance, we mask out select belief modules when computing attention. This experiment is similar to occluding parts of images to identify which
features play a causal role in predictions from classification CNNs~\citep{zeiler_eccv14}.
First, we mask out individual belief modules (\tableref{tab:masking_runs} ``Masked out'')
and find that in general, individual module exclusion minimally affects the agent. In fact, SPL slightly rises when \cpcat1 is excluded. This may indicate these modules provide redundant, generic representations. However, when \cpcat8 is excluded, the agent's performance drops dramatically
below baseline results ($0.233$~\vs$0.712$ on SPL with and without masked \cpcat8).
Surprisingly, correlating attention with actions in~\secref{sec:model_analysis_appendix} reveals little about why \cpcat8 may be special.

The inverse diagnostic, masking all modules except one, suggests that the agent is not relying entirely on \cpcat8.
For example, if we only use \cpcat1, the agent can still reach $0.143$ SPL, though using \cpcat8 reaches a higher $0.225$ SPL.

\begin{figure}
    \centering
    \includegraphics[width=0.95\linewidth]{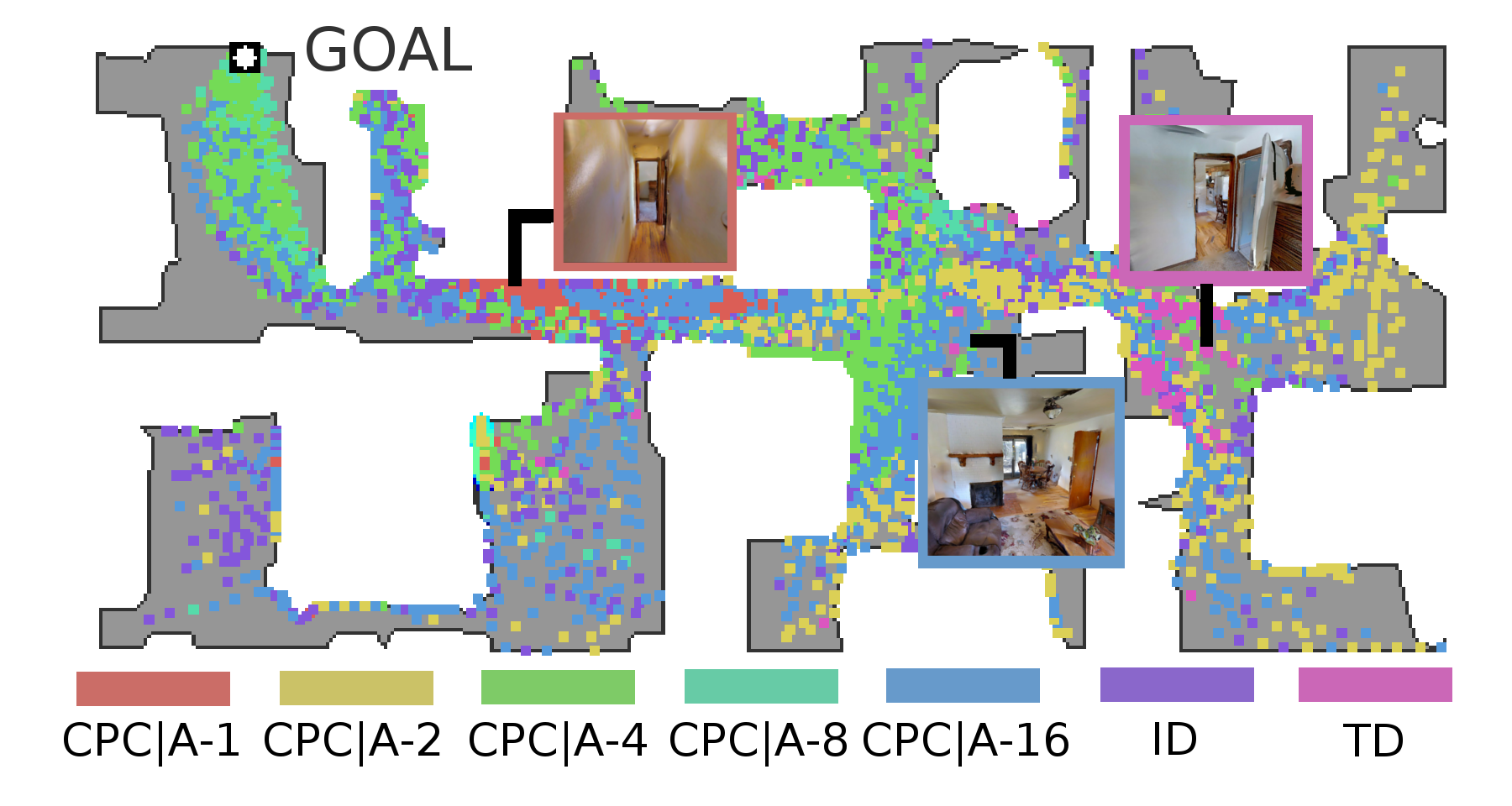}
    \caption{Top-down map of the Cantwell scene in the Gibson dataset~\citep{xia_cvpr18}. Colored boxes represent the auxiliary task with maximum attention at the given location. Attention appears to correlate with agent location.}
    \label{fig:top_down}
\end{figure}

\xhdr{Auxiliary task attention distribution based on location in environment.}
We also qualitatively examine the attention distribution conditioned on location
for an environment from the validation set in~\figref{fig:top_down}. We run this with our most sample efficient model, \allcpc+ID+TD: Attn+E.
We randomly sample $200$ spawn locations with a fixed goal and color-code trajectories according to the auxiliary task
belief module maximally attended to.
Interestingly, the visualization exhibits clustering,
suggesting the agent associates environment patterns with specific belief modules. This location-characterized activation evokes the notion of `place cells' discovered in rats navigating mazes~\citep{burgess_noble14}. We conjecture that these characteristic distributions emerge naturally, analogous to the emergence of specialized kernels in CNNs. Different from transferred specialized representations as in~\citep{shen_iccv19}, we see from-scratch training can learn specialized features. Indeed, we run a variant with separate belief modules without any auxiliary tasks and again observe specialized distributions.

\iftoggle{arxiv}{
    \begin{table*}[h]
        \centering
        \resizebox{.975\textwidth}{!}{
            \begin{tabular}{@{}lcccccccccc@{}}

                && Control & \cpcat1 & \cpcat2 & \cpcat4 & \cpcat8 & \cpcat16 & \cpcat\{1,2\} \\
                \toprule
                 \multirow{2}{*}{\shortstack[l]{Masked\\out}}
                 & Success & 0.911 &
                \textbf{0.416} &	0.815 &	0.901 & 0.878 & 0.906 & 0.058
                \\
                & SPL & 0.777 &
                \textbf{0.224} & 0.711 & 0.800 & 0.772 & 0.793 & 0.012
                \\
                \cdashline{1-11}
                \multirow{2}{*}{\shortstack[l]{All others\\masked out}}
                & Success & 0.911
                & \textbf{0.419} & 0.074 & 0.043 & 0.078 & 0.005 & 0.772
                \\
                & SPL & 0.777 &
                \textbf{0.320} & 0.042 & 0.020 & 0.054 &  0.002 & 0.690
                \\
                \bottomrule
            \end{tabular}
        }
        \caption{Masking \cpcat\{1-16\}: Attn+E modules. \cpcat1, \cpcat2, associated with stopping are critical.}
        \label{tab:masking_runs}
    \end{table*}
}{
    \begin{table*}[t]
        \centering
        \resizebox{.975\textwidth}{!}{
            \begin{tabular}{@{}l|cccccc|ccc@{}}
                \toprule
                Control &
                \multirow{2}{*}{Exclude}
                &
                \cpcat1 & \cpcat2 & \cpcat4 & \cpcat8 & \cpcat16 &
                \multirow{2}{*}{Include}
                & \cpcat{1} &\cpcat{8} \\
                \cline{1-1}
                \cline{3-7}
                \cline{9-10}
                0.712 & &
                0.723 & 0.712 & 0.706 & \textbf{0.233} & 0.706 &

                & 0.143 & 0.225
                \\
                \bottomrule
            \end{tabular}

        }
        \caption{We mask modules in one run of \cpcat\{1-16\}+ID+TD: Attn+E.
        \cpcat8 is critical but not sufficient for performance. Excluding ID, TD omitted for brevity, they perform similarly as \cpcat2.}
        \label{tab:masking_runs}
    \end{table*}
}

\section{Conclusion}
\label{sec:conc}

We have shown that auxiliary tasks can greatly accelerate learning in \pointnav.
We systematically disentangle improvements in performance due to 1) individual
auxiliary tasks, 2) naive combination of multiple auxiliary tasks by summing losses,
and 3) attention over representations from multiple auxiliary tasks, which performs best.
\iftoggle{arxiv}{Our best model achieves $0.61s$ SPL in $8$M observations -- $5\times$ better sample-efficiency over the baseline.
This speedup suggests auxiliary tasks can be key in training
embodied agents in complex environments from scratch within
practical computation budgets.

Our analysis further reveals insightful task and data relationships -- attention extracts information from \allcpc that renders ID and TD unhelpful, even though the tasks provide orthogonal gains when naively combined. Further, attention agents learn to specialize their modules -- \eg specific modules became responsible for driving stopping behavior.
In future work, we aim to study module specialization in more detail, and explore auxiliary objectives
for other embodied tasks (\eg language-driven navigation~\citep{anderson_cvpr18},
question-answering~\citep{embodiedqa},~\etc).
}{Our best model achieves $0.55$ SPL in $7$M observations -- $5.5\times$ better sample-efficiency over the baseline.
This speedup suggests auxiliary tasks can be key in training
embodied agents in complex environments from scratch within
practical computation budgets. Our analysis further reveals agents learn to specialize their modules -- \eg specific modules became responsible for driving stopping behavior.
In future work, we aim to further study this approach for other embodied tasks (\eg language-driven navigation~\citep{anderson_cvpr18},
question-answering~\citep{embodiedqa},~\etc).}

\section{Acknowledgments}
The Georgia Tech effort was supported in part by NSF, AFRL, DARPA, ONR YIPs, ARO PECASE, Amazon.
AD was supported in part by fellowships from Facebook, Adobe, and Snap Inc.
The views and conclusions contained herein are those of the authors and should not be interpreted as necessarily representing the official policies or endorsements, either expressed or implied, of the U.S. Government, or any sponsor.

\newpage
\LetLtxMacro{\section}{\oldsection}
{
\small
\bibliography{strings,main}
}

\newpage
\renewcommand\thesection{\Alph{section}}
\setcounter{section}{0}
\renewcommand\thefigure{A\arabic{figure}}
\renewcommand\thetable{A\arabic{table}}
\setcounter{figure}{0}
\setcounter{table}{0}
\phantomsection

\section{Appendix}

In this supplement we will perform additional analysis, covering an ablation study~\secref{sec:ablative_analysis}, a review of trends in Success~\secref{sec:experiments-appendix}, and model analysis in~\secref{sec:model_analysis_appendix}. We provide plots of our method applied to a harder environment, in~\secref{sec:mp3d}, showing that our main trends still hold. We then consider details omitted in the main paper, describing auxiliary tasks in~\secref{sec:aux_detailed}, fusion methods in~\secref{sec:fusion_details}, and training details in~\secref{sec:training_details}. We conclude with a few additional figures. 
\iftoggle{arxiv}{}{
}
\subsection{Success}
\label{sec:experiments-appendix}
\begin{table*}[t]
    \centering
    \resizebox{0.975\textwidth}{!}{
        \begin{tabular}{@{}lccccc@{}}
            & \multicolumn{2}{c}{Area under Curve} & & \multicolumn{2}{c}{Best} \\
            \cline{2-3} \cline{5-6}
            & Success $(\mathbf{\uparrow})$ & SPL $(\mathbf{\uparrow})$ & & Success $(\mathbf{\uparrow})$ & SPL $(\mathbf{\uparrow})$ \\[0.05in]
            \toprule
\rownumber Baseline & 
 $0.583 $\scriptsize{$\pm 0.061$} & $0.422 $\scriptsize{$\pm 0.043$}
& &
 $0.714 $\scriptsize{$\pm 0.011$} & $0.545 $\scriptsize{$\pm 0.010$}
\\
\rownumber \cpcat$1$ & 
 $0.652 $\scriptsize{$\pm 0.040$} & $0.480 $\scriptsize{$\pm 0.032$}
& &
 $0.796 $\scriptsize{$\pm 0.010$} & $0.628 $\scriptsize{$\pm 0.011$}
\\
\rownumber \cpcat$2$ & 
 $0.654 $\scriptsize{$\pm 0.040$} & $0.487 $\scriptsize{$\pm 0.035$}
& &
 $0.797 $\scriptsize{$\pm 0.011$} & $0.641 $\scriptsize{$\pm 0.008$}
\\
\rownumber \cpcat$4$ & 
 $0.680 $\scriptsize{$\pm 0.030$} & $0.512 $\scriptsize{$\pm 0.029$}
& &
 \boldsymbol{$0.815 $}\scriptsize{\boldsymbol{$\pm 0.015$}} & $0.658 $\scriptsize{$\pm 0.014$}
\\
\rownumber \cpcat$8$ & 
 $0.681 $\scriptsize{$\pm 0.034$} & $0.517 $\scriptsize{$\pm 0.027$}
& &
 \boldsymbol{$0.815 $}\scriptsize{\boldsymbol{$\pm 0.007$}} & \boldsymbol{$0.658 $}\scriptsize{\boldsymbol{$\pm 0.012$}}
\\
\rownumber \cpcat$16$ & 
 $0.677 $\scriptsize{$\pm 0.039$} & $0.514 $\scriptsize{$\pm 0.029$}
& &
 \boldsymbol{$0.819 $}\scriptsize{\boldsymbol{$\pm 0.019$}} & \boldsymbol{$0.663 $}\scriptsize{\boldsymbol{$\pm 0.010$}}
\\
\rownumber ID & 
 $0.655 $\scriptsize{$\pm 0.052$} & $0.458 $\scriptsize{$\pm 0.036$}
& &
 $0.798 $\scriptsize{$\pm 0.013$} & $0.588 $\scriptsize{$\pm 0.011$}
\\
\rownumber TD & 
 $0.612 $\scriptsize{$\pm 0.062$} & $0.441 $\scriptsize{$\pm 0.044$}
& &
 $0.756 $\scriptsize{$\pm 0.017$} & $0.564 $\scriptsize{$\pm 0.017$}
            \\[0.01in]
            \cdashline{1-6}
            
            \rownumber \allcpc: Add & 
 $0.665 $\scriptsize{$\pm 0.028$} & $0.523 $\scriptsize{$\pm 0.026$}
& &
 $0.831 $\scriptsize{$\pm 0.007$} & \boldsymbol{$0.687 $}\scriptsize{\boldsymbol{$\pm 0.010$}}
\\
\rownumber \allcpc+ID+TD: Add & 
 $0.682 $\scriptsize{$\pm 0.035$} & $0.532 $\scriptsize{$\pm 0.028$}
& &
 \boldsymbol{$0.842 $}\scriptsize{\boldsymbol{$\pm 0.008$}} & \boldsymbol{$0.696 $}\scriptsize{\boldsymbol{$\pm 0.013$}}

            \\[0.01in]
            \cdashline{1-6}
            
            \rownumber \allcpc+ID+TD: Average & 
 $0.678 $\scriptsize{$\pm 0.026$} & $0.539 $\scriptsize{$\pm 0.022$}
& &
 $0.851 $\scriptsize{$\pm 0.003$} & $0.696 $\scriptsize{$\pm 0.008$}
\\
\rownumber \allcpc+ID+TD: Softmax\cite{shen_iccv19} & 
 $0.725 $\scriptsize{$\pm 0.022$} & $0.578 $\scriptsize{$\pm 0.020$}
& &
 $0.838 $\scriptsize{$\pm 0.008$} & $0.696 $\scriptsize{$\pm 0.008$}
\\

\rownumber \allcpc+ID+TD: Attn & 
 $0.724 $\scriptsize{$\pm 0.031$} & $0.565 $\scriptsize{$\pm 0.027$}
& &
 \boldsymbol{$0.856 $}\scriptsize{\boldsymbol{$\pm 0.016$}} & $0.698 $\scriptsize{$\pm 0.015$}
\\
\rownumber \allcpc+ID+TD: Attn+E & 
 $0.756 $\scriptsize{$\pm 0.019$} & $0.594 $\scriptsize{$\pm 0.019$}
& &
 \boldsymbol{$0.854 $}\scriptsize{\boldsymbol{$\pm 0.008$}} & \boldsymbol{$0.707 $}\scriptsize{\boldsymbol{$\pm 0.006$}}
\\

            \cdashline{1-6}
            
            \rownumber Depth (Fine-tuned) & 
 $0.584 $\scriptsize{$\pm 0.042$} & $0.454 $\scriptsize{$\pm 0.037$}
& &
 $0.773 $\scriptsize{$\pm 0.020$} & $0.612 $\scriptsize{$\pm 0.012$}
\\
\rownumber Depth (Frozen) & 
 $0.583 $\scriptsize{$\pm 0.051$} & $0.451 $\scriptsize{$\pm 0.045$}
& &
 $0.772 $\scriptsize{$\pm 0.040$} & $0.616 $\scriptsize{$\pm 0.033$}
\\
            \bottomrule
        \end{tabular}
    }
    \caption{Primary quantitative results summarized. We show 1. individual auxiliary tasks greatly improve \pointnav learning efficiency, (rows 2-8) 2. adding multiple tasks naively yields marginal gains, and (rows 9-10) 3. separating task influences into separate modules recovers early learning penalties and further improves learning. (row 11-14). Bolded variants dominate their group (denoted with dashed lines).}
    \label{tab:results-all_appendix}
\end{table*}
Success shows similar trends as SPL, as shown by the results in~\tableref{tab:results-all_appendix}. We attach success plots reproduced with SPL plots as additional figures. In the following, we provide some observations.
\begin{compactitem}[--]
    \item TD gains 4\% and ID gains 10\% in success over the baseline (rows 8, 7, 1) by 40M frames, a larger difference than in SPL. This shows the metrics don't strictly trend together. ID and TD agents wander more but still are better able to reach the goal than the baseline. 
    \item The slight SPL edge that \cpc-\{4,8,16\} have over \cpc-\{1,2\} is still reflected in Success.
    \item Success is highly similar among agents in rows 9-14. This is expected, as \pointnav learning is sharply logarithmic~\cite{wijmans_iclr20}. It takes much longer to improve as the agent matures, and so a small speedup won't move metrics much. 
    \item Depth agents (rows 15, 16) still do not exceed \cpcat1 in success. 
\end{compactitem}

\iftoggle{arxiv}{}{

}

\subsection{Auxiliary Task Details}
\label{sec:aux_detailed}
\begin{figure}[t]
    \centering
    \includegraphics[width=0.95\textwidth]{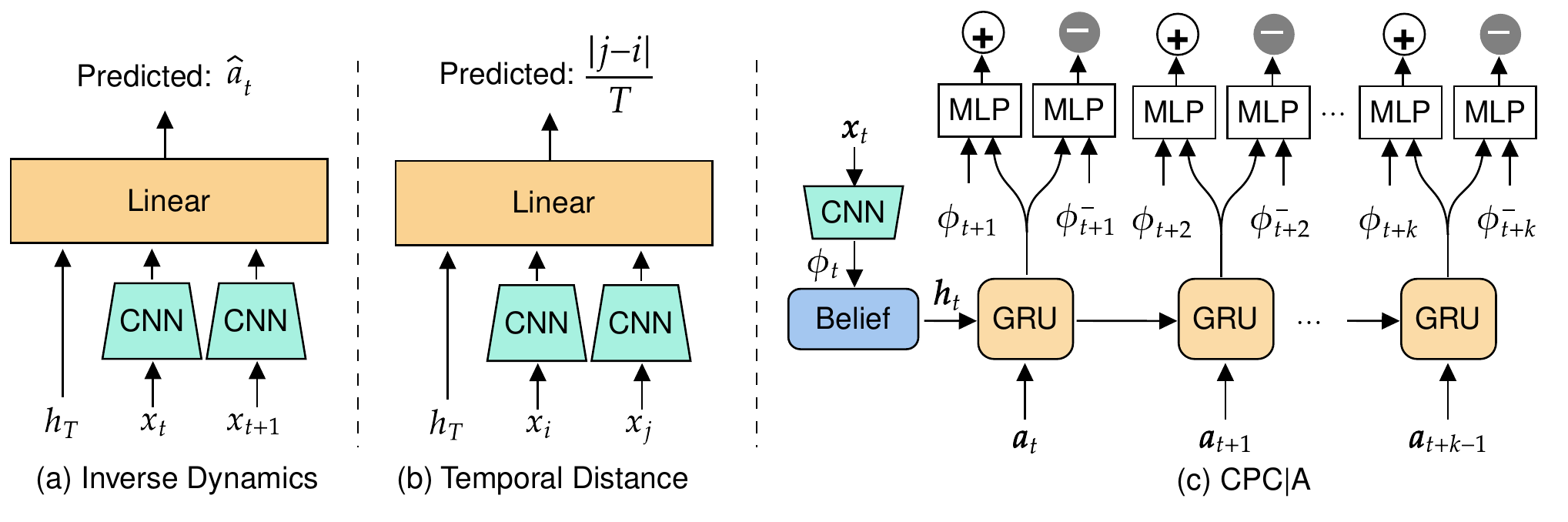}
    \caption{We study three auxiliary modules (re-printed for reference).
    a) Inverse dynamics: decoding action taken from successive visual embeddings
    $\phi_t$ and $\phi_{t+1}$ and the final belief state $h_T$.
    b) Temporal distance: decoding the timestep difference between two observation
    embeddings from final belief state $h_T$.
    c) \cpc: decoding future observation embeddings ($\phi_{t+1}, ...,  \phi_{t+k}$)
        at every timestep from other observation embeddings using a secondary GRU.}
    \label{fig:id-td-cpc_appendix}
\end{figure}

Here, we elaborate on the computation done in each auxiliary module.

\xhdr{Inverse Dynamics (ID).}
As shown in~\figref{fig:id-td-cpc_appendix}a, given two successive
observations ($x_t$ and $x_{t+1}$) and the belief module hidden state at the
end of the trajectory ($h_T$), the ID task is to predict the action taken at time
$t$, $a_t$. We include the belief module hidden state to encourage representation
of trajectory actions in it.

Specifically, we take the visual embeddings from $1-T$, trim the final timestep to form the "before" batch, and the first timestep to from the "after batch. We then concatenate each timestep $t$ and $t+1$ pair with the belief module output from timestep $t$, and predict action logits. We use cross-entropy loss with the true actions from timesteps $1$ to $(T-1)$, and subsample the loss by 0.1.
\begin{align}
    L_{ID} &= \sum_{i=1}^{T-1}L_{CE}(I(\phi_{i}, \phi_{i+1}, h_T), a_i)
\end{align}

\xhdr{Temporal Distance (TD).}
As shown in~\figref{fig:id-td-cpc_appendix}b, given two observations from a trajectory ($x_i$ and $x_j$) and
the belief module hidden state at the end of the trajectory ($h_T$),
the TD task is to predict $\frac{|j-i|}{T}$.
This is similar in spirit to progress estimation in~\citep{ma_iclr19} and
reachability in~\citep{savinov_iclr18}.
However, rather than Euclidean or geodesic distance, we ask the agent to predict the
(normalized) number of steps elapsed between two visual observations.
This requires the agent to recall if a location is revisited or similarly viewed,
designed to promote understanding of spatio-temporal relations of trajectory viewpoints.

In detail, we select $k=8$ random pairs of indices, and get their corresponding visual embeddings. We concatenate the pairs' visual embeddings with the belief module's end output, (\ie $h_T$), and directly use a linear layer to predict the timestep difference between each pair.
\begin{align}
    L_{TD} &= \frac{1}{2}((i - j) - \mathcal{T}(\phi_{i}, \phi_{j}, h_T))^2
\end{align}

\xhdr{Action-Conditional Contrastive Predictive Coding (\cpc)}.
As shown in~\figref{fig:id-td-cpc_appendix}c,
given the belief module hidden state $h_t$, a second GRU is unrolled for $k$ timesteps using future actions $\{a_{t+i}\}_{i=0}^{k-1}$ as input.
The output of the second GRU at time $t+i$ is used to distinguish different visual representations.
We concatenate the second GRU's output at $t+i$ with a) the ground-truth visual representation at $t+i$, $\phi_{t+i+1}$, or b) a ``negative'' visual feature $\phi_{t+i+1}^-$ sampled from other timesteps and trajectories. Then, "contrasting" the different representations can be framed as a classification task, classifying inputs with the ground-truth visual representation as $1$, and negatives as $0$. This encourages $h_t$ to build long-horizon representations of the environment.

Technically, a \cpc-specific GRU with hidden size 512 is initialized with output $h_t$ from the belief module. Its $k$ input actions are first fed through a size 4 embedding layer. The \cpc GRU then outputs $g_1^t, g_2^t, \dots, g_k^t$. These outputs are concatenated with positive and negative visual embeddings, and fed into a two layer decoder (hidden size 32). The decoder predicts logits for whether the input contained a positive or negative visual embedding, which is fed into a cross entropy loss given targets 1 and 0 respectively. We perform this for all timesteps $1$ to $(T-k)$, and subsample the loss by 0.2.

\begin{align}
    L_{\cpc} &= \sum_{k=1}^K\sum_{t=1}^{T-k} L_{CE}(c(g_{k}^t, \phi^-), 0) + L_{CE}(c(g_{k}^t, \phi_{t+k}), 1)
\end{align}

\subsection{Module Fusion Details}
\label{sec:fusion_details}

\begin{figure}[t]
    \centering
    \includegraphics[width=12cm]{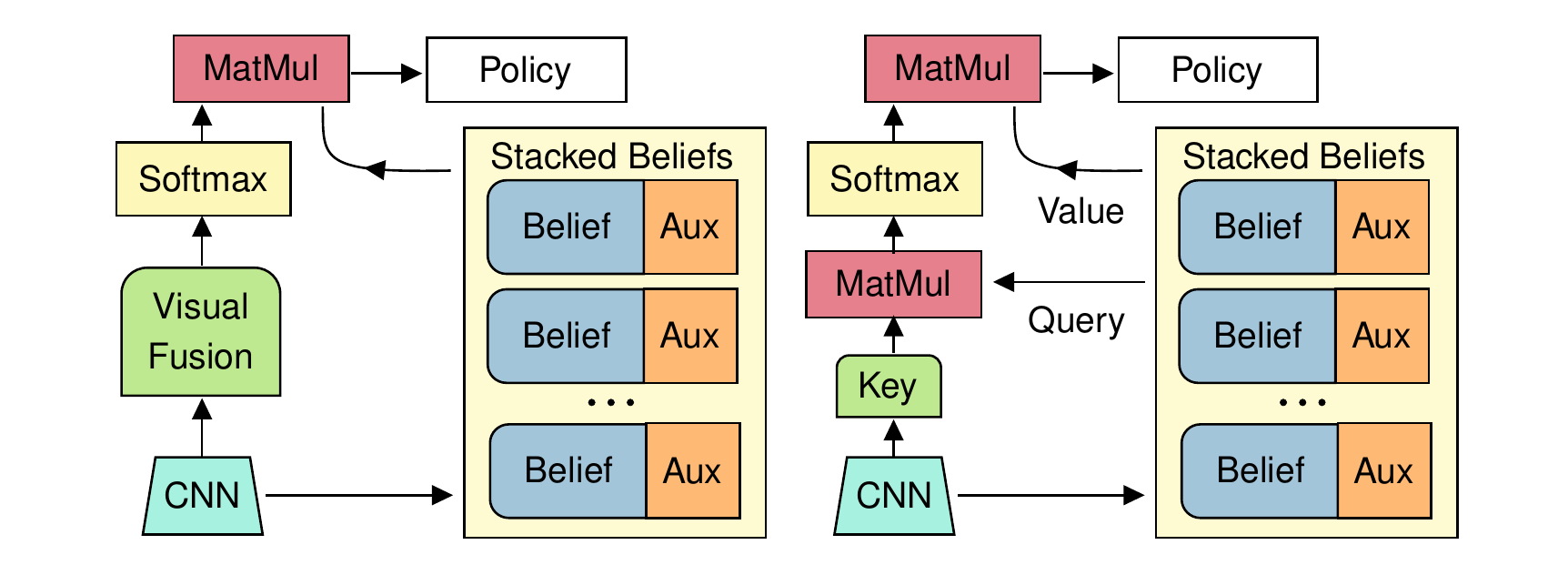}
    \caption{Left: Dot-product attention, Right: Softmax gating.}
    \label{fig:supp_fusion}
\end{figure}

All fusion methods are achieved by a form of weighted sum. Fixed and average fusion are achieved by freezing weights as desired. Weight calculation for softmax gating and attention are described by \figref{fig:supp_fusion}\iftoggle{arxiv}{, re-printed for convenience}{}. In softmax gating, a linear layer directly converts the visual embedding into logits that are passed through a softmax layer to create weights. With dot-product attention, the visual representation is passed through a "key" linear layer, outputting a key of size 512. The representations given by the separate belief modules serve as queries, which are multiplied by the key to create our logits. These logits are again put through a softmax layer to create our final weights.

\iftoggle{arxiv}{}{
\xhdr{Entropy}
We use a variant of scaled dot-product attention
with an entropy penalty (denoted `\textsc{+E}'). Given attention distribution
$w_{\text{attn}}\vcentcolon= (p_1,\dots,p_{n_{\text{Aux}}})$, we calculate entropy
as $\sum_{i=1}^{n_{\text{Aux}}} -p_i\log p_i$. Entropy encourages the agent to
use multiple belief modules. An agent that quickly learns to use a single module may prevent the other modules from learning about the task (from reduced gradients). 
}

\subsection{More Model Analysis}
\label{sec:model_analysis_appendix}

\begin{figure}
\centering
\begin{tabular}{m{0.18\textwidth}| m{0.18\textwidth}m{0.18\textwidth}m{0.18\textwidth}m{0.18\textwidth}}
\includegraphics[width=0.17\textwidth]{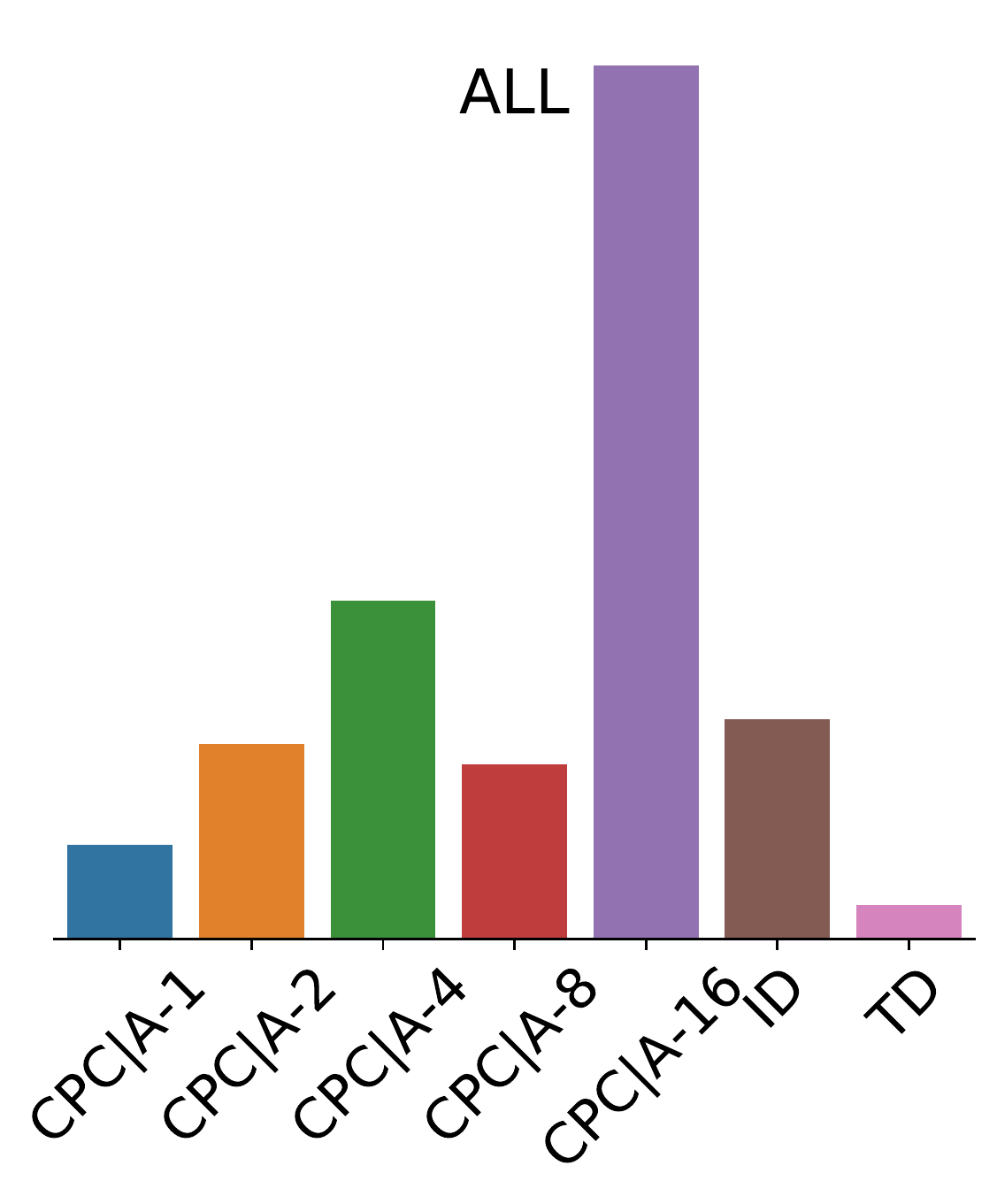} &
\includegraphics[width=0.17\textwidth]{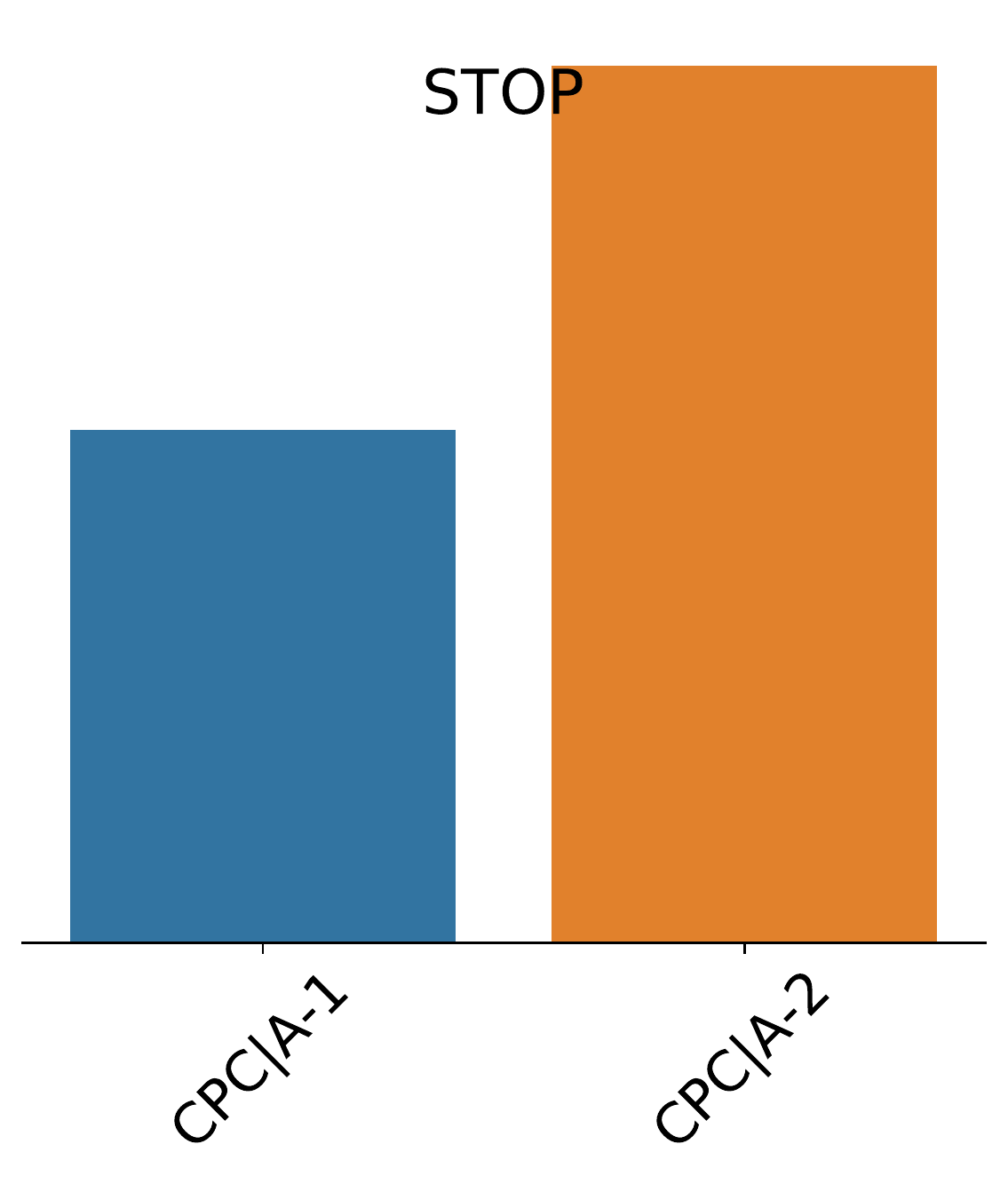} &
\includegraphics[width=0.17\textwidth]{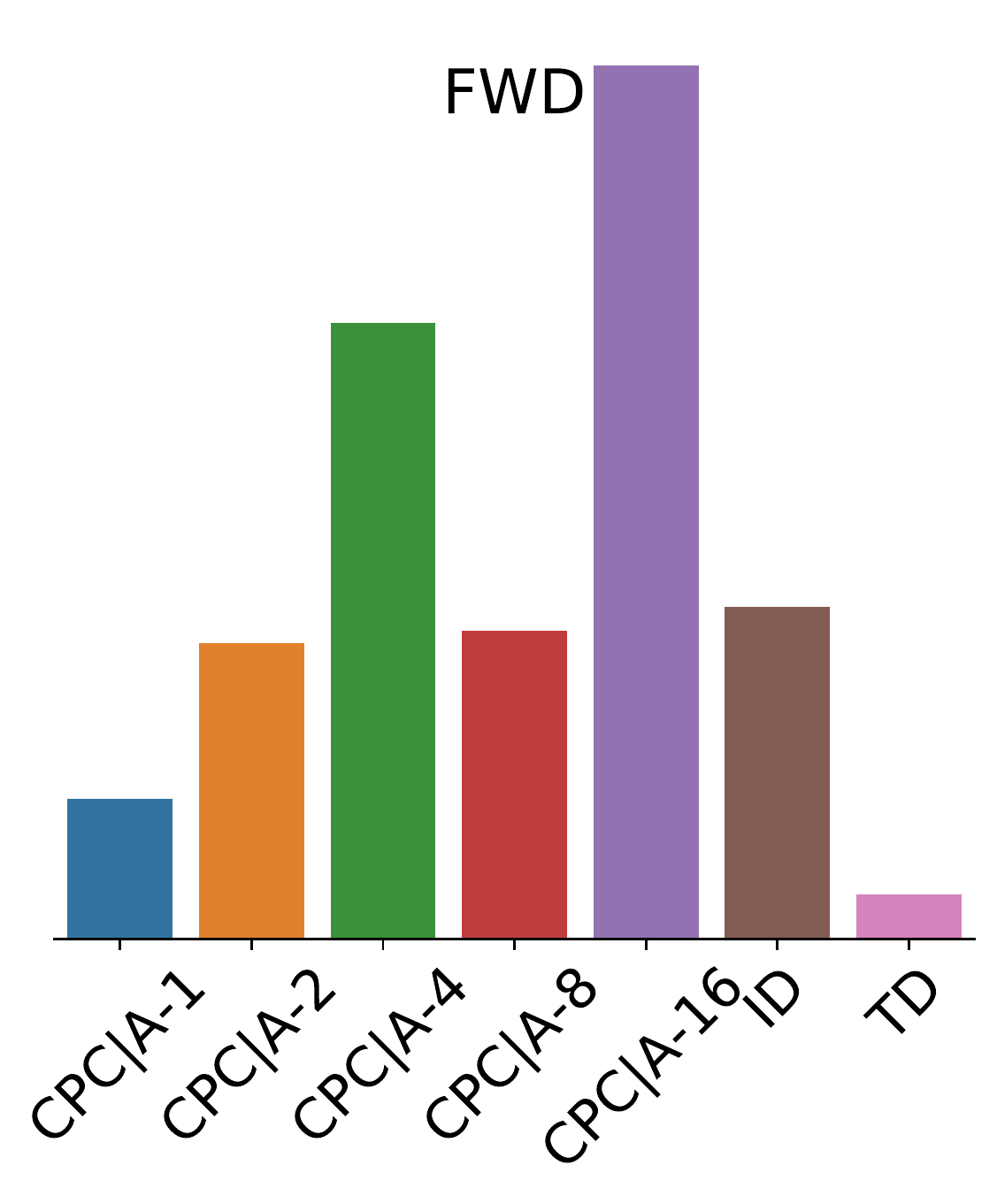} &
\includegraphics[width=0.17\textwidth]{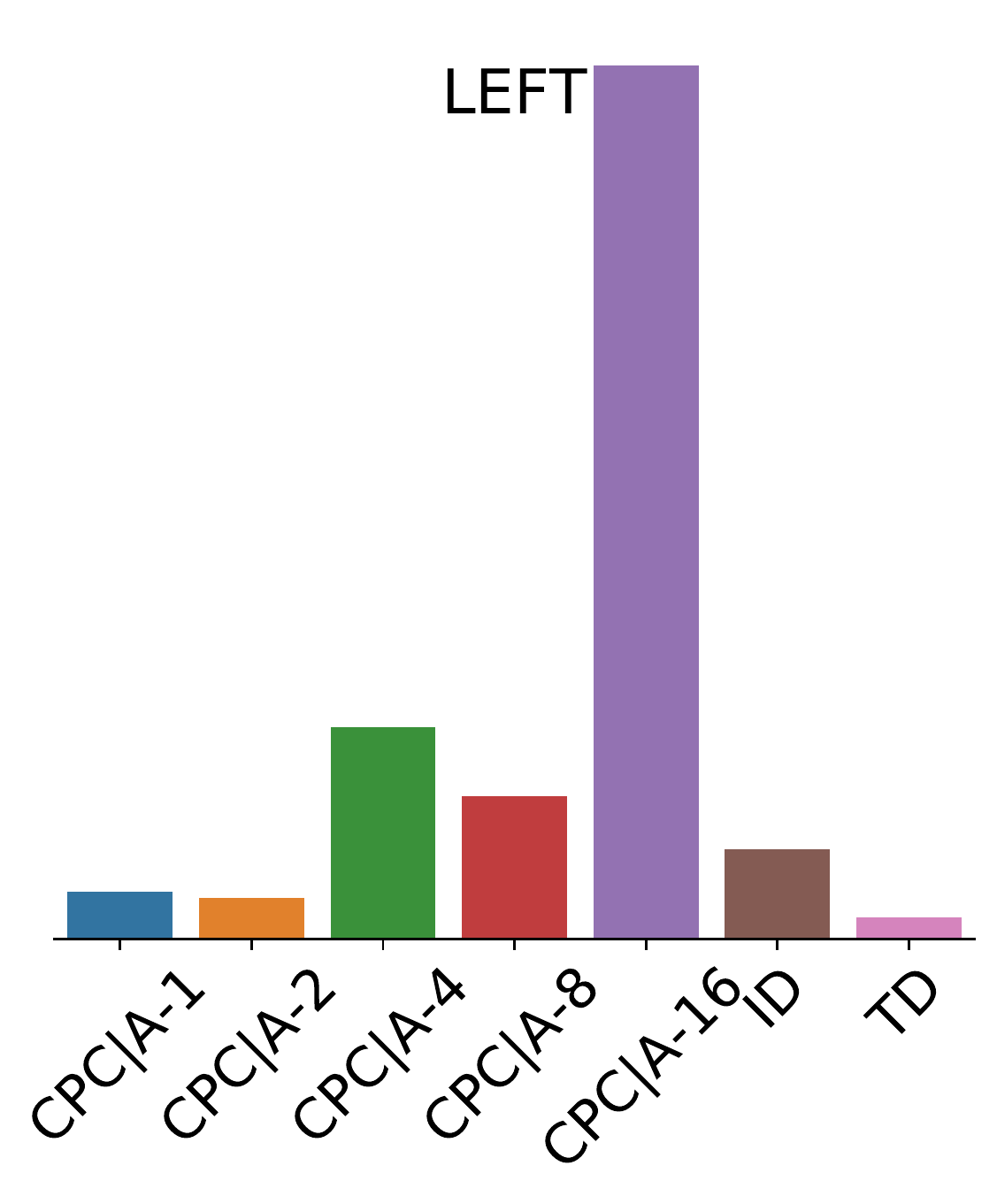} &
\includegraphics[width=0.17\textwidth]{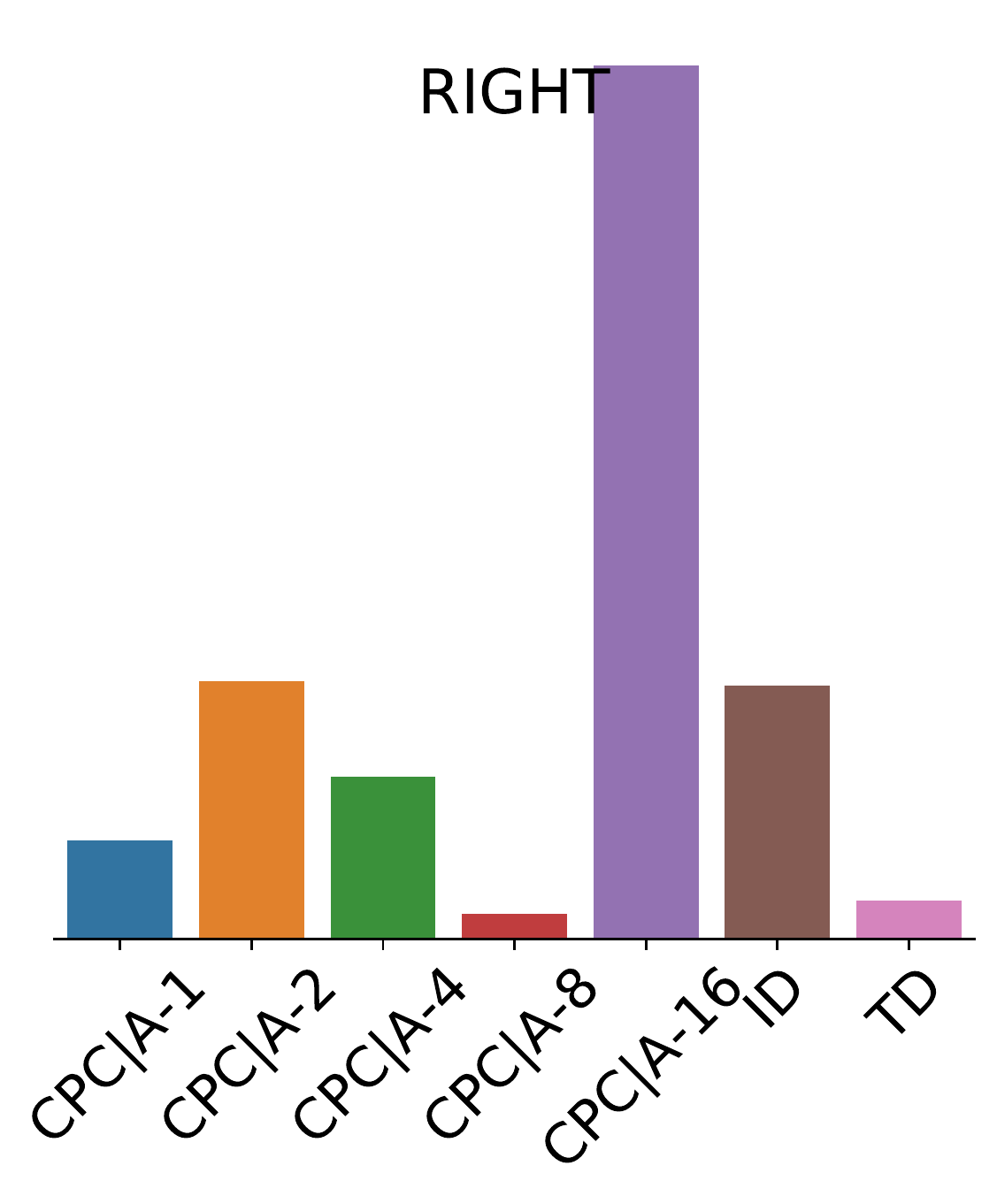}
\end{tabular}
\caption{Distribution of which auxiliary tasks are most attended to while taking each action for
    \allcpc: Attn+E. \cpcat1 and \cpcat2 significantly
    affect the \textsc{stop} action}
\label{fig:distributions}
\end{figure}

\begin{figure}
\centering
\begin{tabular}{m{0.3\textwidth}m{0.3\textwidth}m{0.3\textwidth}}
\includegraphics[width=0.25\textwidth]{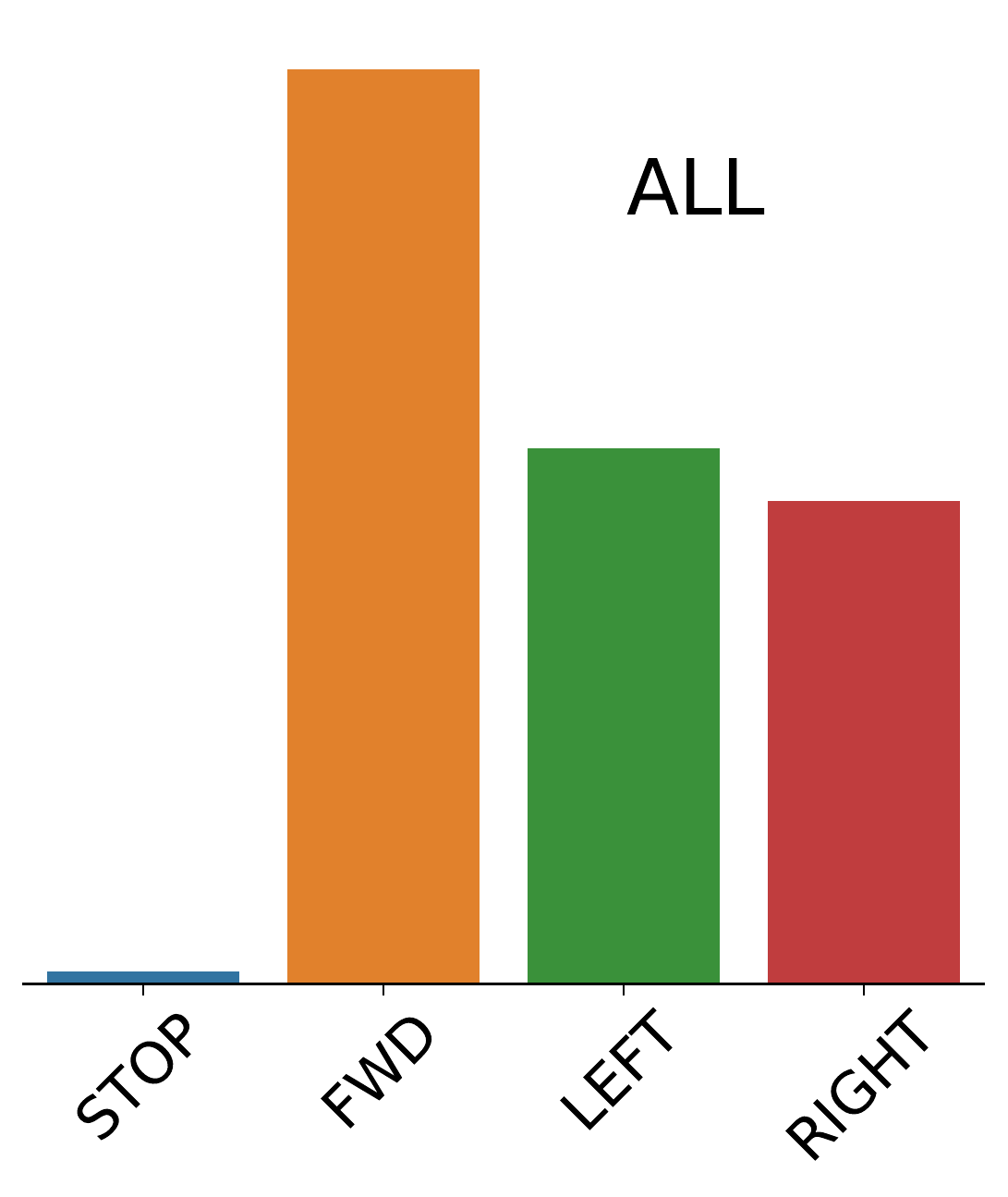} &
\includegraphics[width=0.25\textwidth]{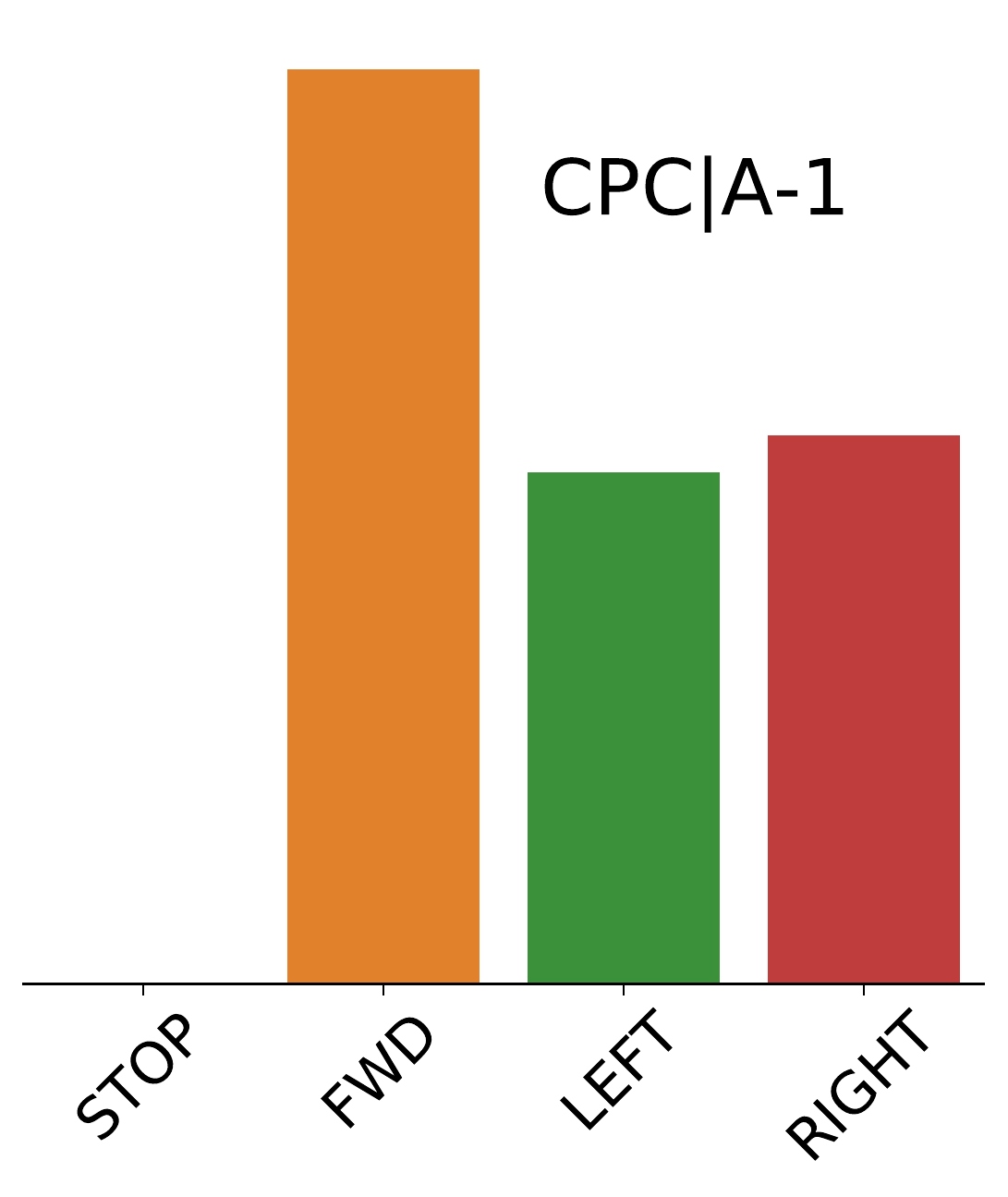} &
\includegraphics[width=0.25\textwidth]{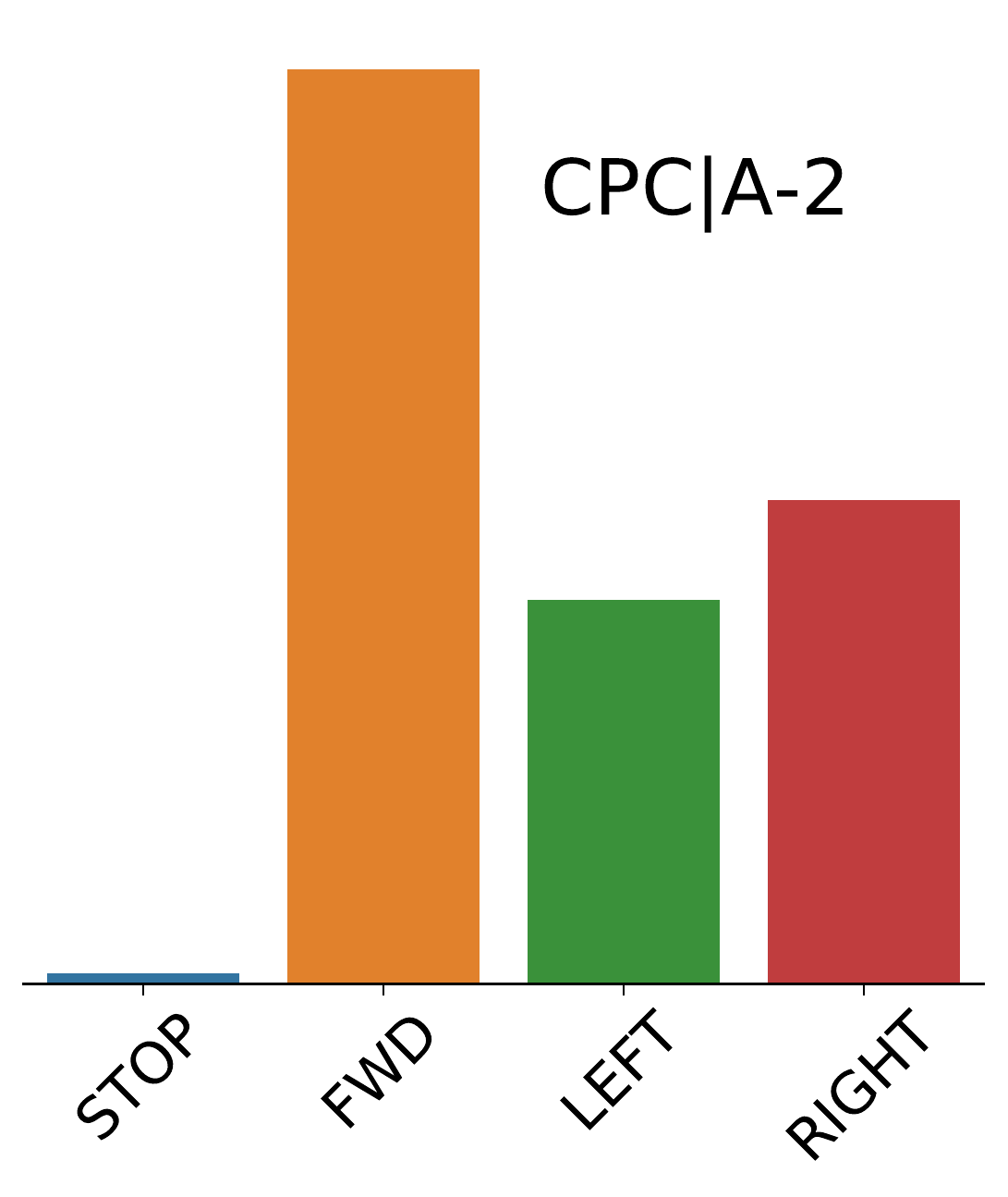}
\\
\includegraphics[width=0.25\textwidth]{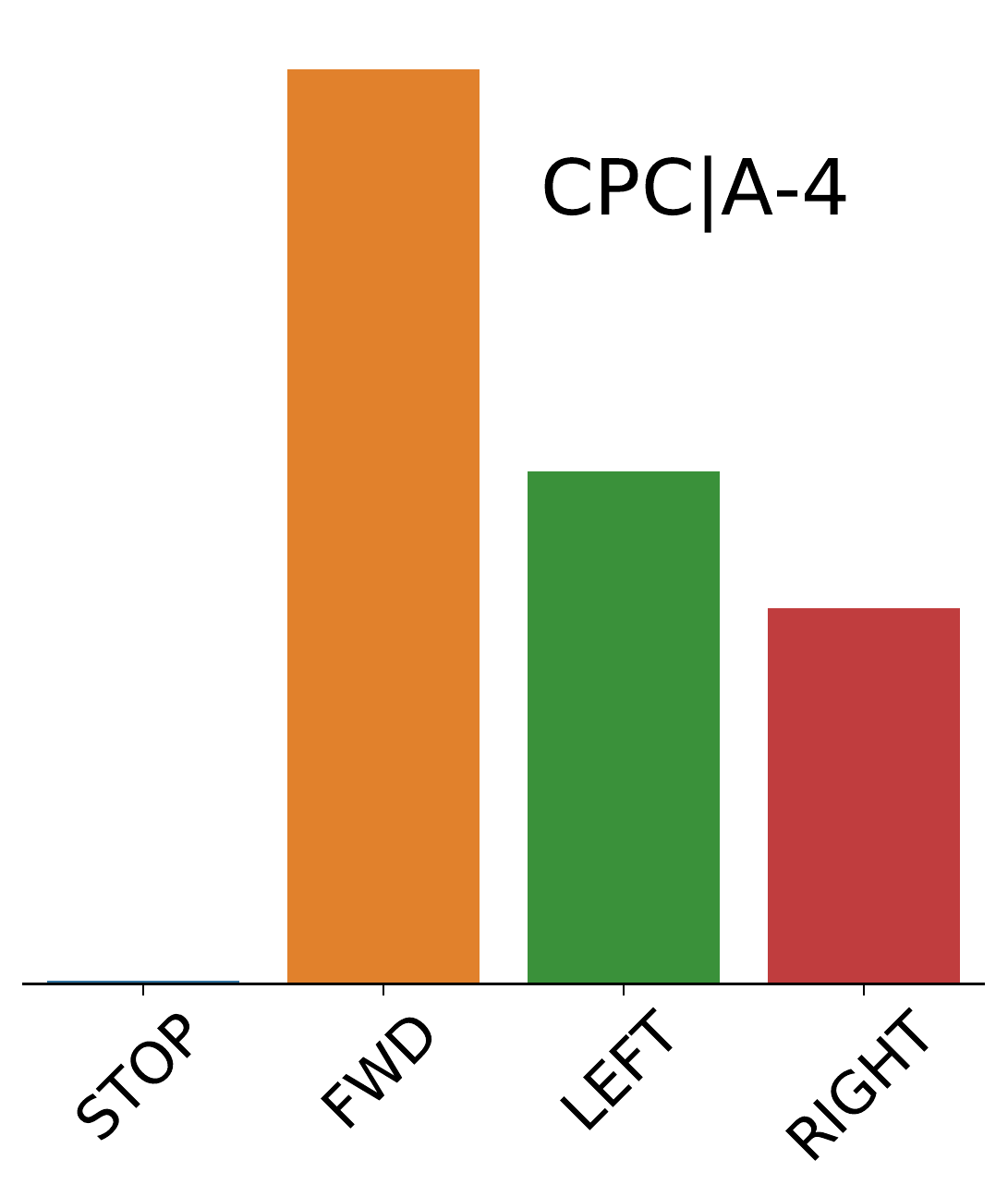} &
\includegraphics[width=0.25\textwidth]{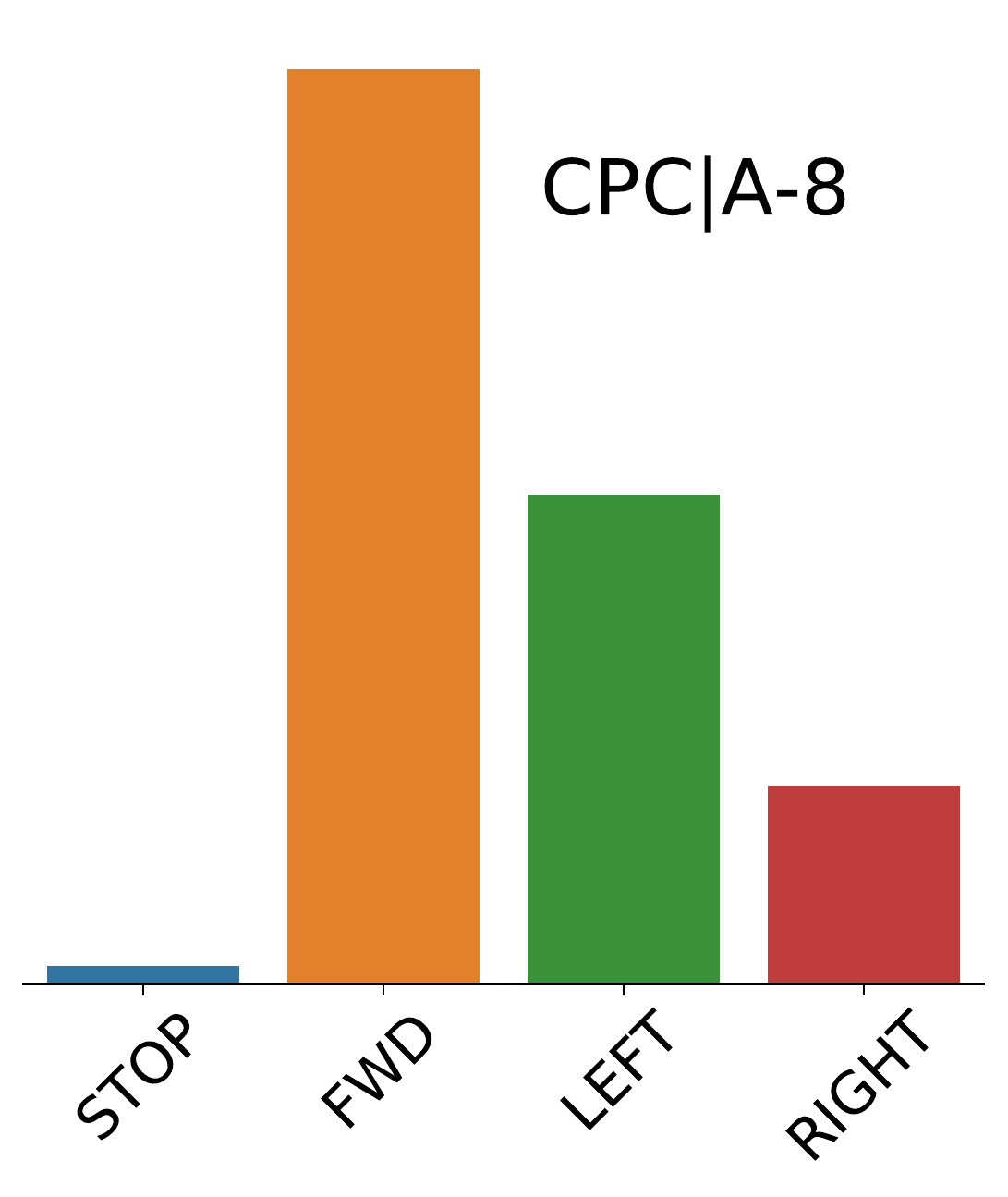} &
\includegraphics[width=0.25\textwidth]{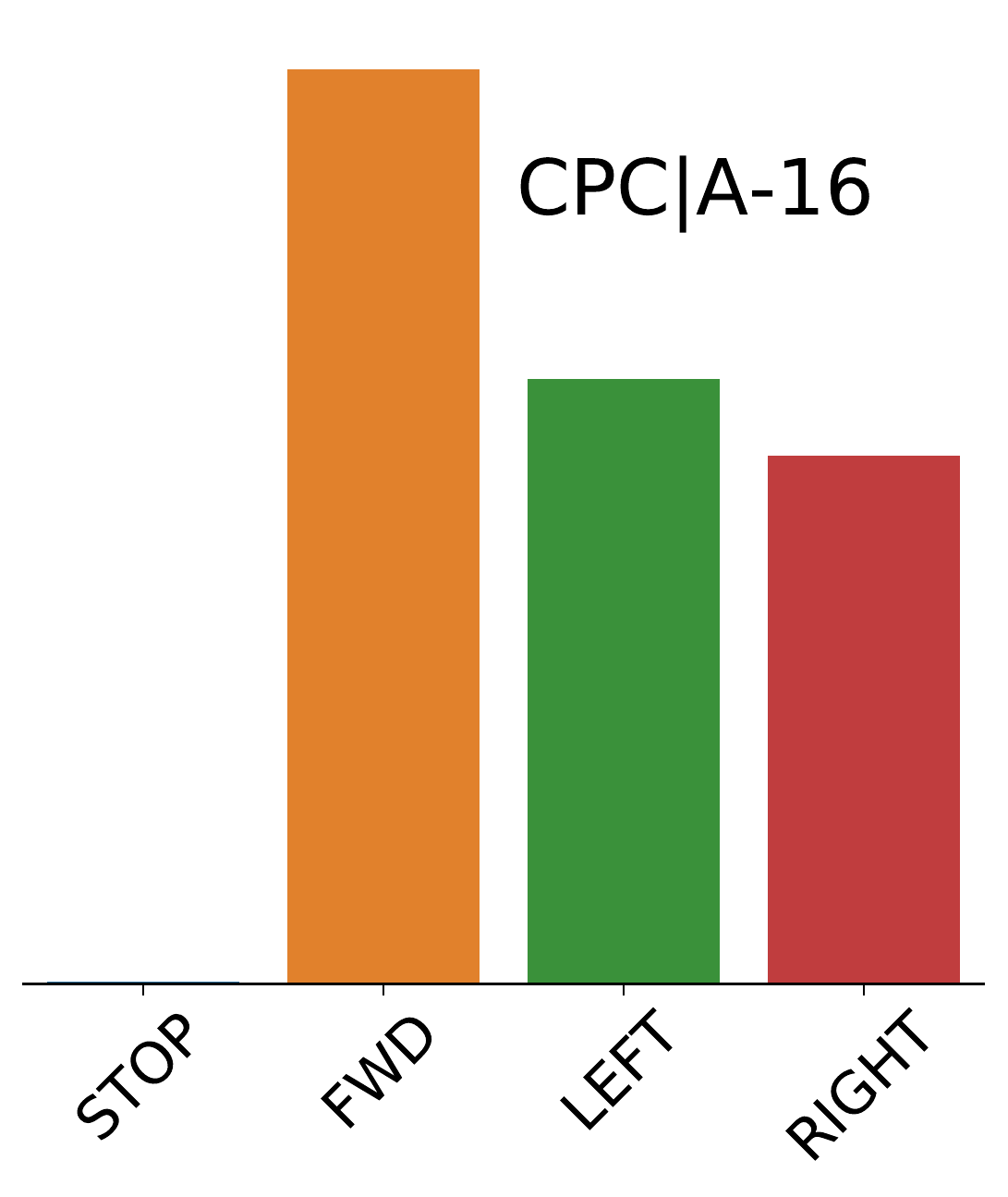}
\\
\includegraphics[width=0.25\textwidth]{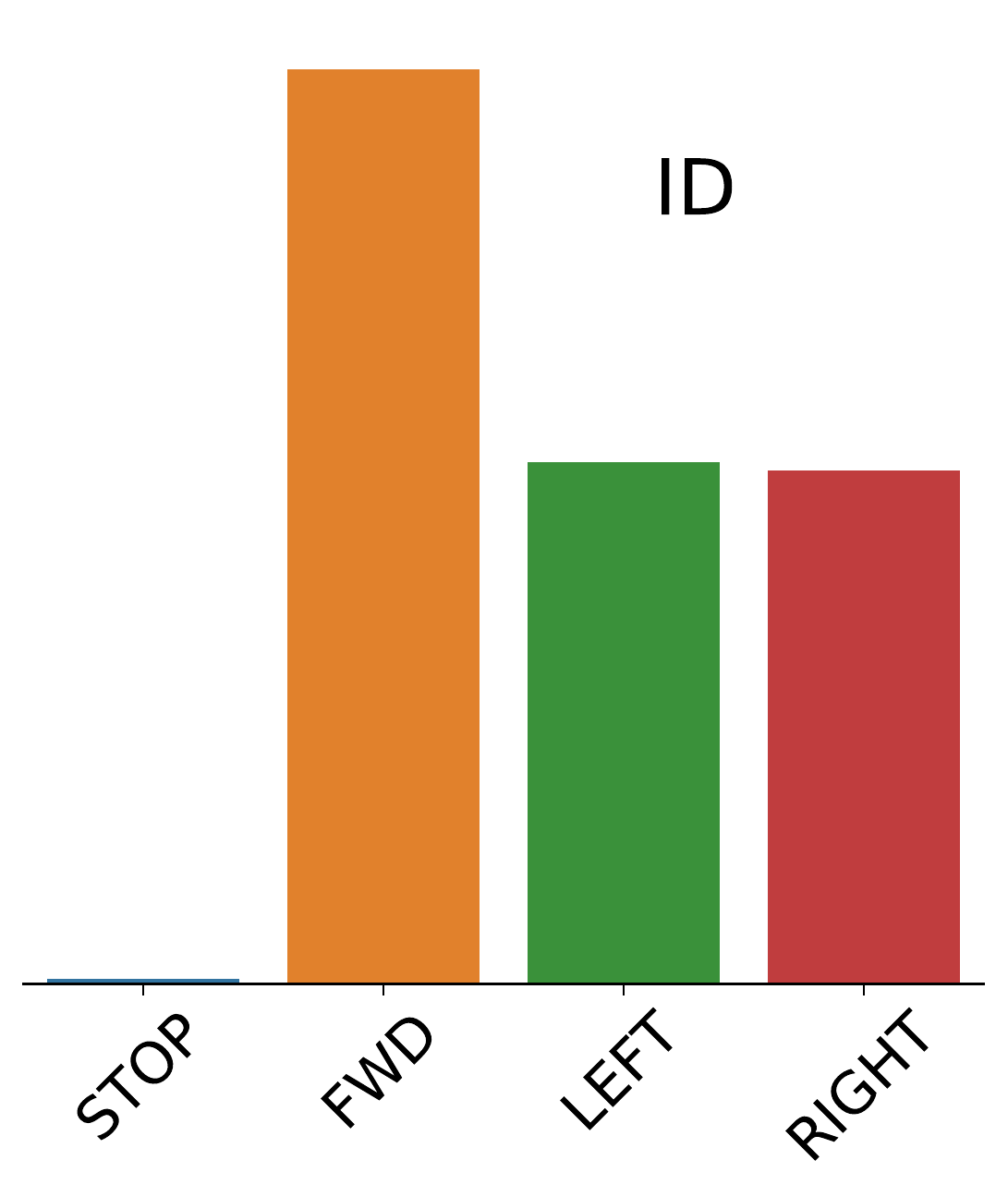} &
\includegraphics[width=0.25\textwidth]{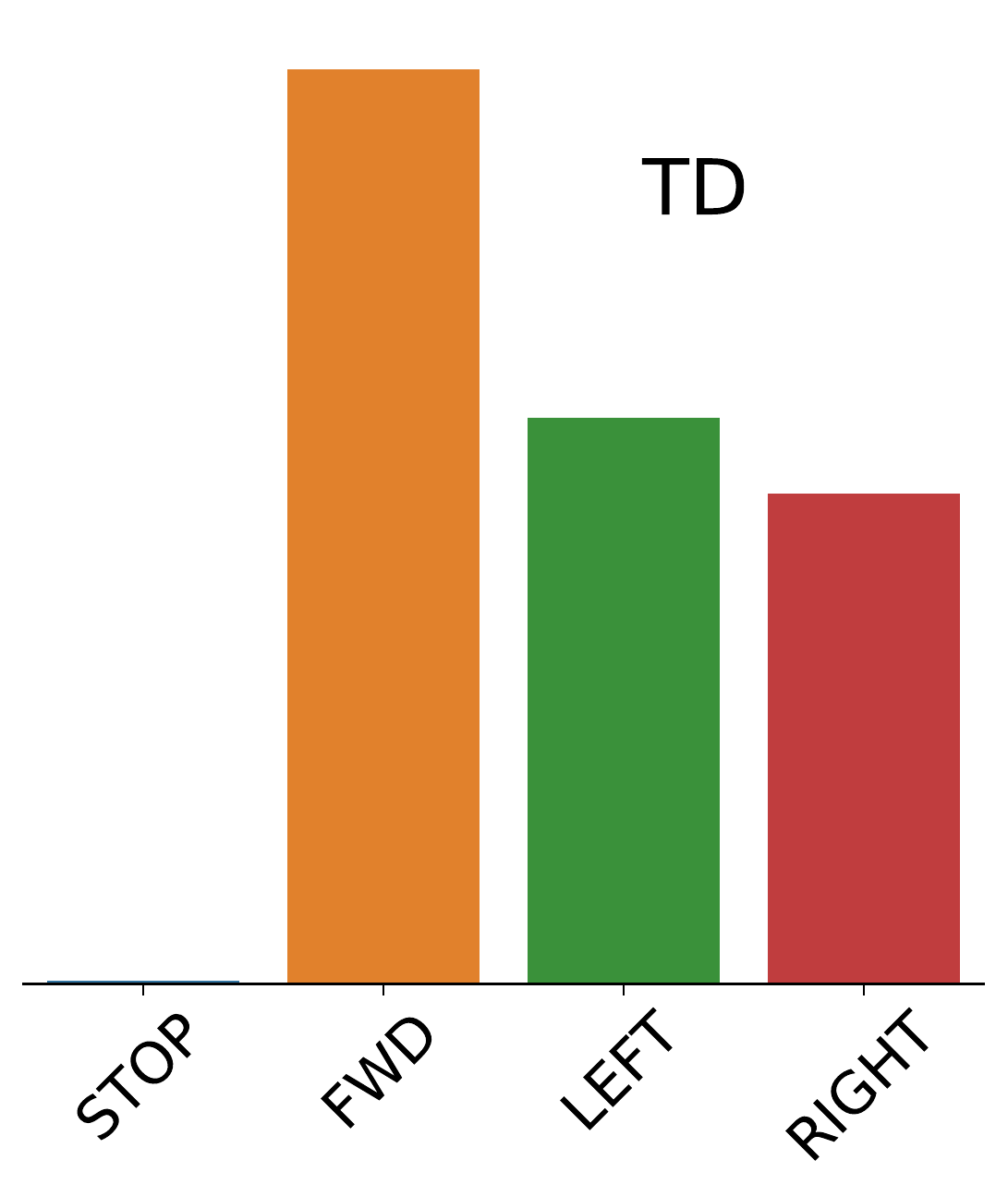}
\\
\end{tabular}
\caption{Action distributions on \allcpc: Attn+E validation episodes, for steps where a given auxiliary task is used (weight $> 0.25$). Again, the \textsc{stop} action reliably activates \cpcat1, \cpcat2.}
\label{fig:supp_action_dist}
\end{figure}

\figref{fig:distributions} shows how auxiliary task attention correlates with the action taken for the same run analyzed in~\secref{tab:masking_runs}.
To compute this, for all agent trajectories in the validation set, we assign credit of +1 to a given action in which an auxiliary task's belief module receives the most attention.
\textsc{ALL} shows the overall count distribution regardless of action taken. In this run, \cpcat16 is attended very frequently, and is highly correlated with turning actions. The \textsc{stop} action almost always corresponds with attention to \cpcat1 and \cpcat2, which may suggest they play an important role in short-term decisions. However, applying the same analysis to other runs of the same variant, we find \emph{different attended modules for \textsc{stop}} (\eg corresponding to \cpcat4). Attention over different belief modules does suggest some functional expertise in \pointnav, but auxiliary tasks do not consistently determine the expertise. 

\xhdr{Auxiliary Tasks have Characteristic Action Distributions}
Instead of conditioning the task distribution on the action as in the main text, we can condition the action distribution on the task. We generate these plots (\figref{fig:supp_action_dist}) by thresholding all actions taken with conditioned task weight $> 0.25$. In this analysis, it is clear that stopping is infrequent enough that \cpcat1, \cpcat2 don't appear particularly associated with the action. Overall, these preliminary analyses suggests attention provides only a shallow explanation for agent behavior.

\subsection{MP3D experiments}
\label{sec:mp3d}

\begin{figure}[t]
    \centering
    \includegraphics[width=6cm]{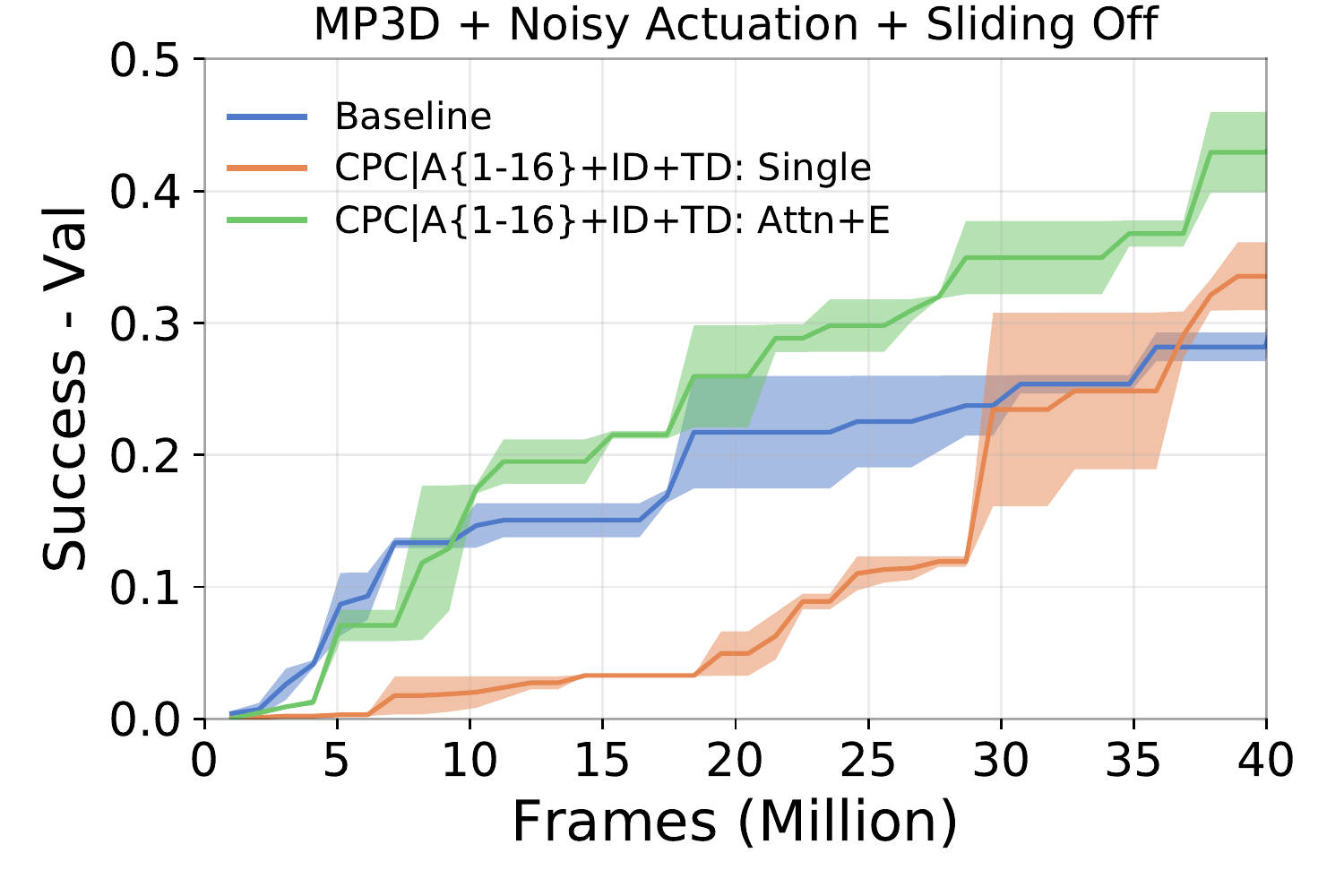}
    \includegraphics[width=6cm]{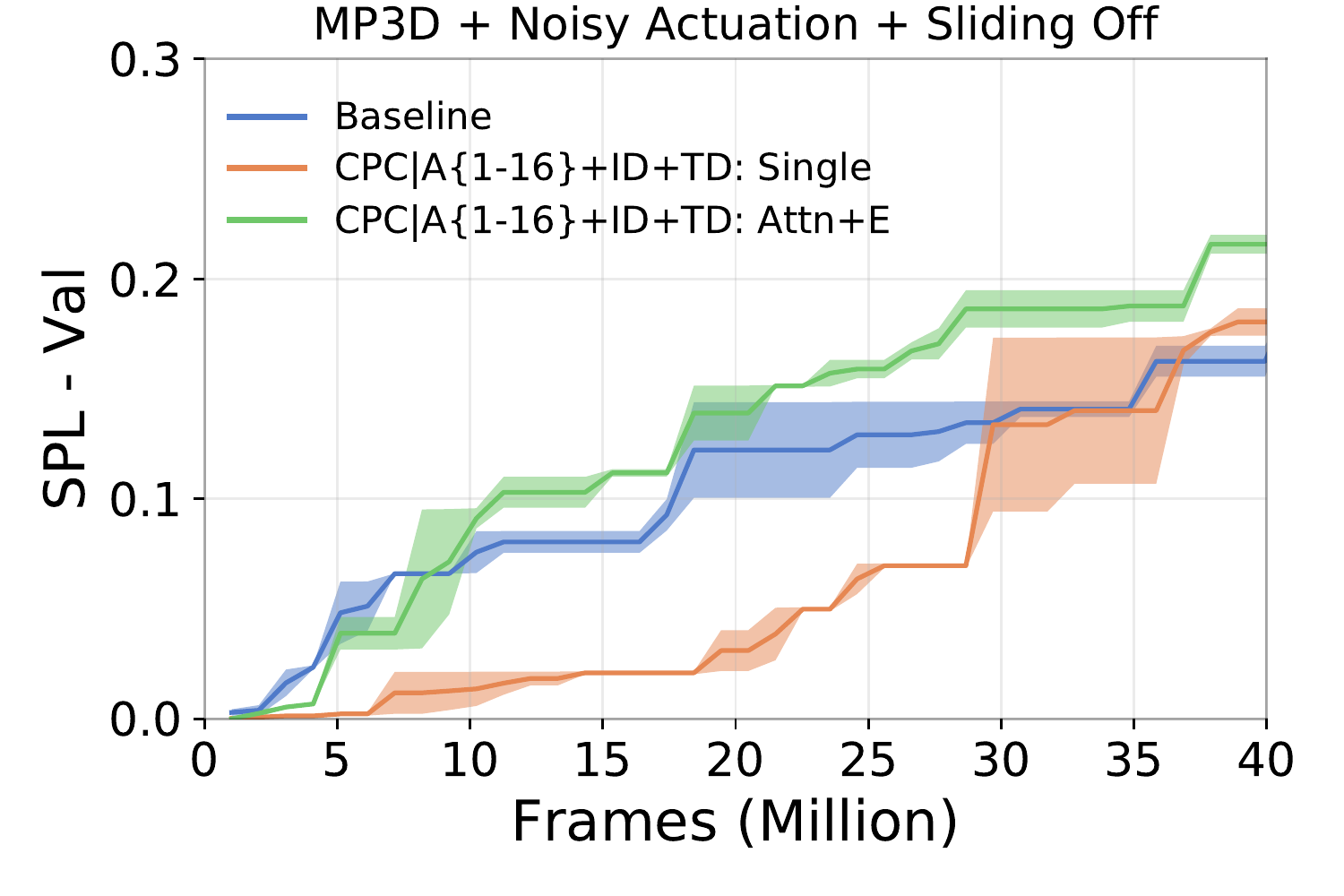}
    \caption{Our method improves on the baseline in Matterport3D.}
    \label{fig:supp_mp3d}
\end{figure}

We run our 3 representative variants on a harder environment to show our method can generalize, showing results in~\figref{fig:supp_mp3d}. Namely, we test combining the auxiliary tasks on one module (Single) and our attentive architecture against the baseline. Our new environment is on the Matterport3D dataset~\citep{mp3d}, with actuation noise and wall sliding turned off, making navigation much more difficult~\cite{kadian2019making}. Nonetheless, using our attentive architecture improves on the baseline, whereas naive application takes much longer to overtake the baseline.

\subsection{Improving Local Planners in Hierarchical Navigation}
\label{sec:local_appendix}
\begin{figure}[h!]
  \centering
  \includegraphics[width=\linewidth]{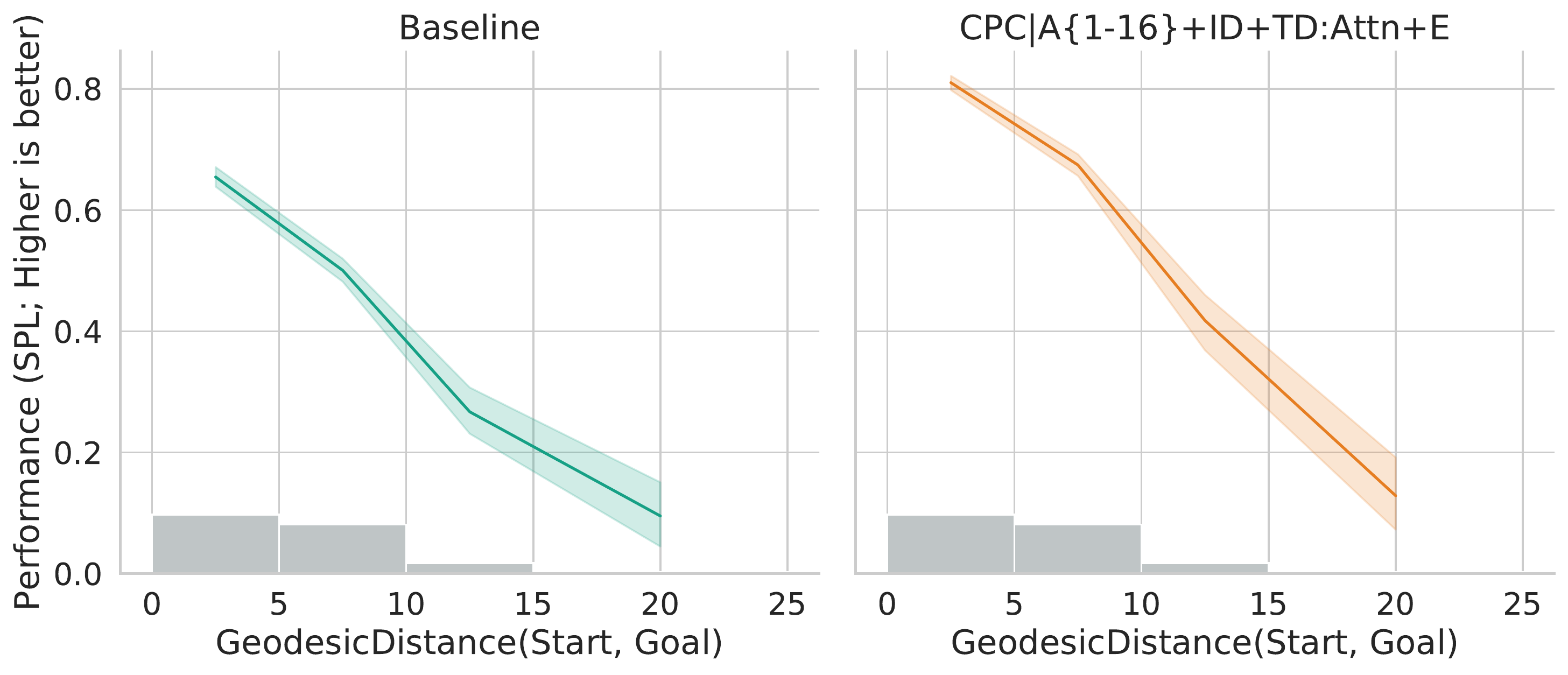}
  \caption{Our method improves navigation at all ranges, thus enabling improvement of local navigation.}
  \label{fig:local_plots_appendix}
\end{figure}

We briefly check that our method improves local navigation \ie improves success and SPL when navigation goals are nearby. We plot SPL with respect to geodesic distance to goal at spawn in \figref{fig:local_plots_appendix}. 
Our method improves metrics across the board, with a slight bias towards improving shorter episodes. Thus, our approach could be adopted in improving local planners in hierarchical agents.

\subsection{Training Details}
\iftoggle{arxiv}{}{
\xhdr{Training.}  We train our agent via Proximal Policy Optimization
PPO)~\citep{schulman_arxiv17} with Generalized Advantage Estimation
(GAE)~\citep{schulman_iclr16}. We use 4 rollout workers with rollout length $T=128$,
and 4 epochs of PPO with 2 mini-batches per epoch.
We set discount factor to $\gamma=0.99$ and GAE factor $\tau=0.95$.
We use the Adam optimizer~\citep{kingma_iclr15} with a learning rate of $2.5\times10^{-4}$ and $\epsilon=0.1$.s
We follow the reward structure in~\citep{savva_iccv19}. For goal $g$,
when the agent is in state $s_t$ and executes action $a_t$ (transitioning to $s_{t+1}$),
\begin{equation}
    r_t(s_t, a_t) = \begin{cases}
    2.5 \cdot \text{Success} & \text{if } a_t = \texttt{stop} \\
    \texttt{GeoDist}(s_{t}, g) -
    \texttt{GeoDist}(s_{t+1}, g) - \lambda & \text{otherwise}
    \end{cases}
\label{eq:pnreward}
\end{equation}
where \texttt{GeoDist} is the geodesic distance and
$\lambda ($=$0.01)$ is a slack penalty.
}
\label{sec:training_details}
No hyperparameter sweeps were done. Model sizes were all $5.7\pm 0.3$ million parameters. This count comprises the visual encoder and the policy networks, but not the decoder networks, though the hidden size is shared throughout the modules. To achieve the uniform model size, single module variants used a GRU with hidden size $512$, while multiple module networks had hidden sizes correspondingly reduced to $\approx 256-288$. 

To evaluate models in t-tests, we select the checkpoints with highest average validation metrics (over 3 validation runs) across 4 training seeds.

The belief module receives input of size 514, 512 from the ResNet-18 visual module, and 2 from the GPS-Compass sensor.

Our complete loss is:
\begin{align}
    L_{\text{total}}(\theta_m;\theta_a) & = L_\text{RL}(\theta_m) - \alpha H_{action}(\theta) + L_\text{Aux}(\theta_m;\theta_a)
    \\ 
    L_\text{Aux}(\theta_m;\theta_a) &= \sum_{i=1}^{n_\text{Aux}}\beta^{i}L^i_\text{Aux}(\theta_m;\theta_a^i) - \mu H_{attn}(\theta_m) 
\end{align}

\begin{figure}[t]
    \centering
    \includegraphics[width=10cm]{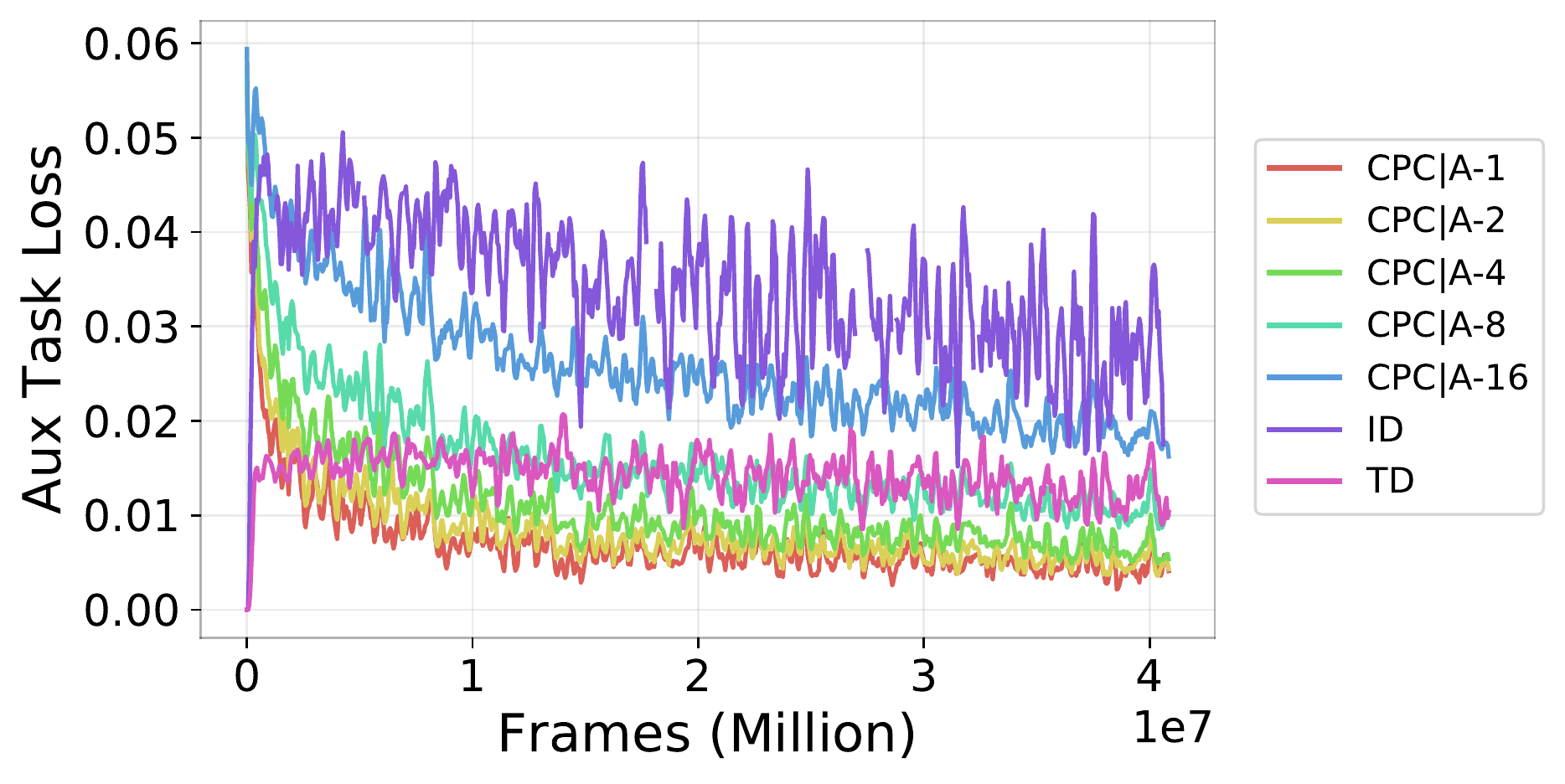}
    \caption{Auxiliary Task loss curves for \cpcat{1-16}+ID+TD: Attn+E.}
    \label{fig:supp_loss}
\end{figure}

$H_{attn}$ is the entropy of the attention distribution over the different auxiliary tasks. In our experiments, we set $\alpha = 0.01$, and $\mu = 0.01$. We set $\beta^i$ for ID and \cpc tasks at $0.1$, and $0.4$ for TD. These values were determined such that the loss terms were in the same order of magnitude at initialization. Losses for the auxiliary tasks trend stably downward as shown in \figref{fig:supp_loss}. The agent does not initially have trajectories sufficiently long (\ie predicting \texttt{stop} after a few steps) to appropriately calculate TD and \cpc tasks, so they start with 0 loss. Other training hyperparameters (some repeated from the main text) are as follows: 
\begin{align}
    \text{Rollout Workers: } n &= 4 \\
    \text{Rollout Length: } t &=128 \\
    \text {PPO Epochs} &= 4 \\  
    \text {PPO Mini-batches} &= 2 \\
    \gamma &= 0.99 \\
    \tau &= 0.95 \\
    \epsilon &= 0.1 \\
    \text{lr} &= 2.5 \times 10^{-4} \\
    \text{Gradient Norm Cap} &= 0.5 \\
    \text{PPO Clip} &= 0.1 \\
\end{align}

\subsection{Additional Figures}

\xhdr{Success and SPL Validation Curves}
We provide the Success validation curves and reproduce the SPL validation curves for reference.

\begin{figure}[h!]
  \centering
  \begin{subfigure}{.5\textwidth}
    \centering
    \includegraphics[width=\linewidth]{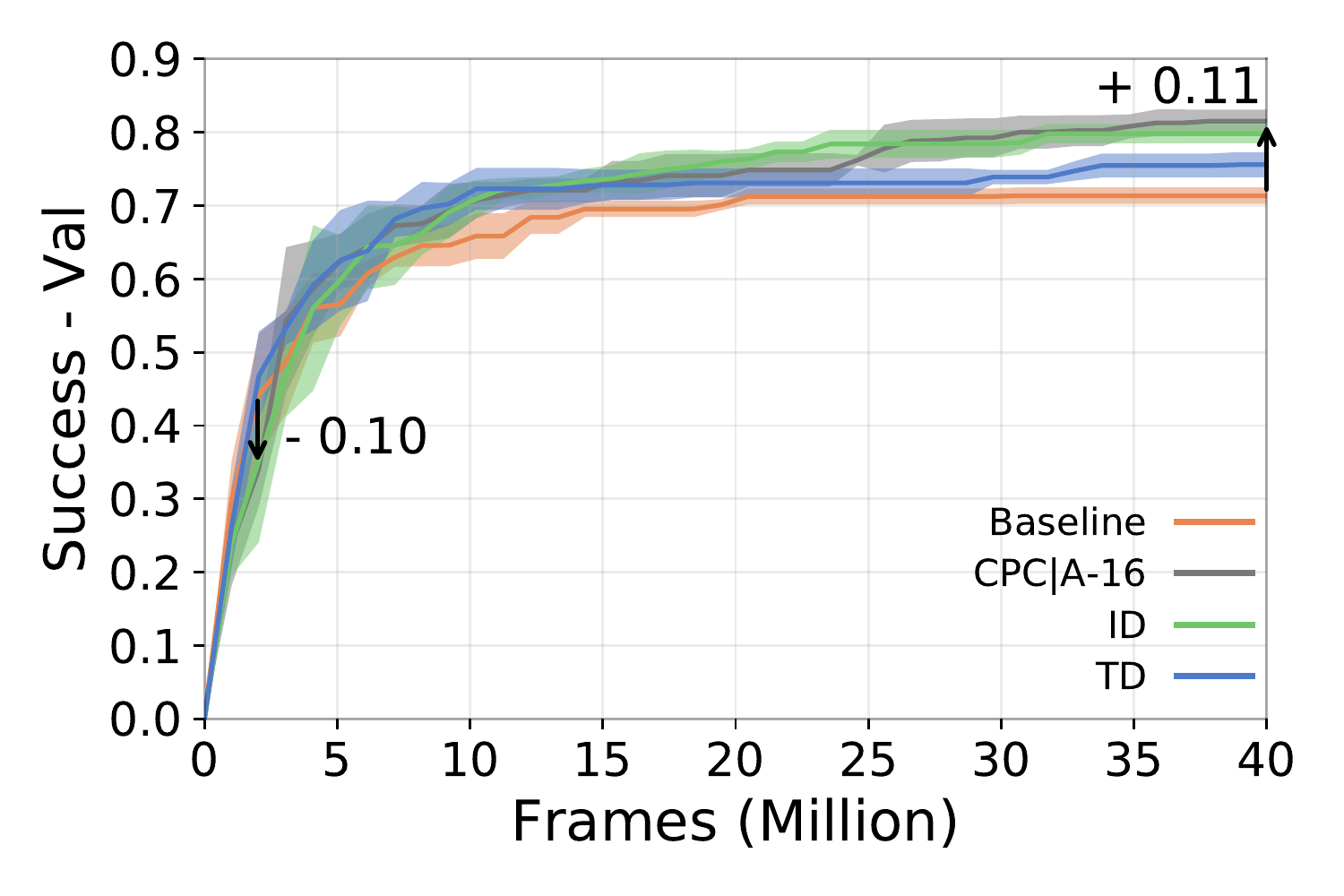}
  \end{subfigure}%
  \begin{subfigure}{.5\textwidth}
    \centering
    \includegraphics[width=\linewidth]{figures/simple_spl.pdf}
  \end{subfigure}
  \caption{All auxiliary tasks overtake the baseline after 5M frames. \cpcat4 provides larger gain than other auxiliary tasks.}
  \label{fig:simple_plots_appendix}
\end{figure}

\begin{figure}[t!]
    \centering
    \begin{subfigure}{.5\textwidth}
      \centering
      \includegraphics[width=\linewidth]{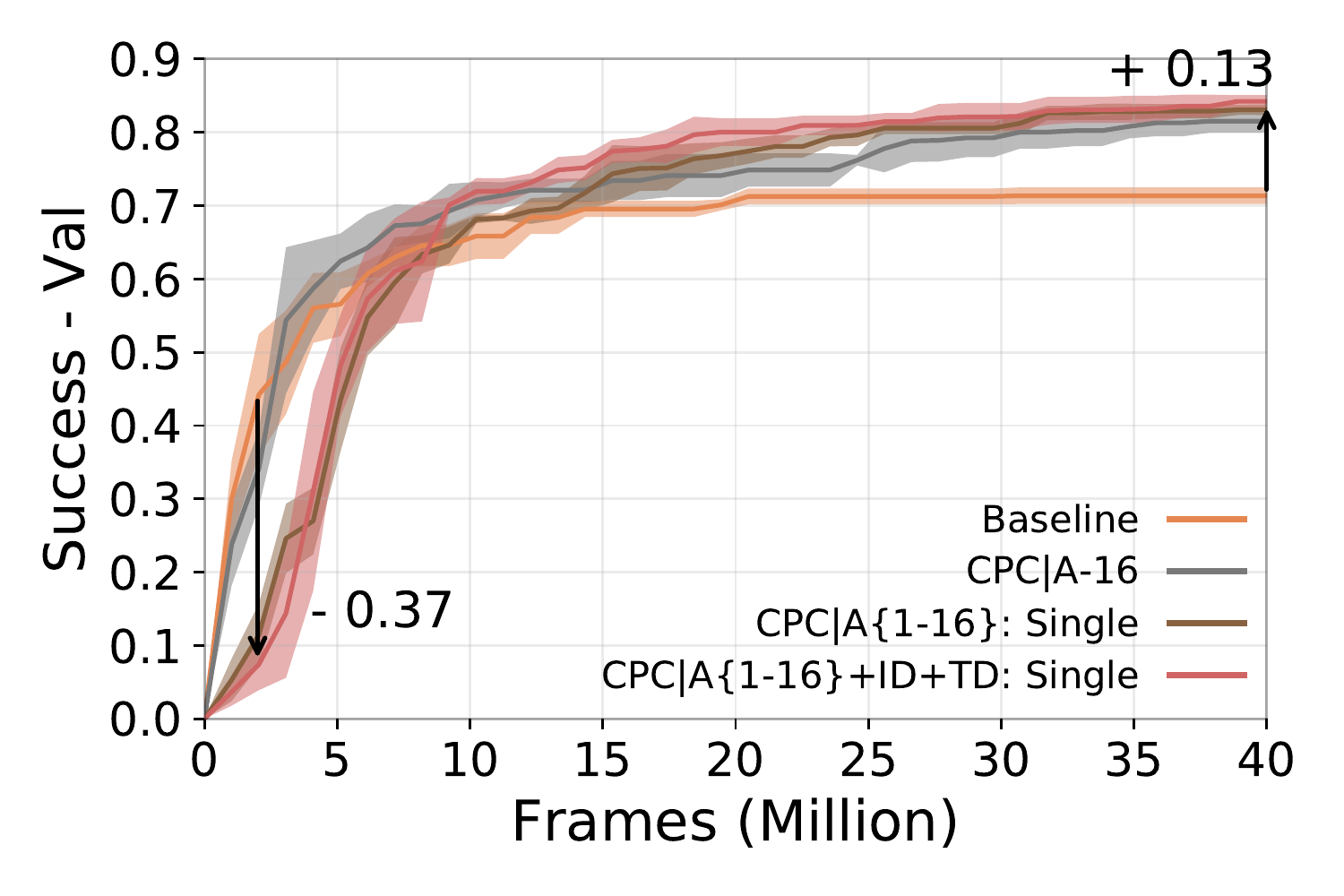}
    \end{subfigure}%
    \begin{subfigure}{.5\textwidth}
      \centering
      \includegraphics[width=\linewidth]{figures/homo_spl.pdf}
    \end{subfigure}
    \caption{Multiple auxiliary task combinations learn better than a single task as agent matures (20-40M frames) at some cost to initial ramp-up.}
    \label{fig:five_plots_appendix}
\end{figure}

\begin{figure}[t!]
    \centering
    \begin{subfigure}{.5\textwidth}
      \centering
      \includegraphics[width=\linewidth]{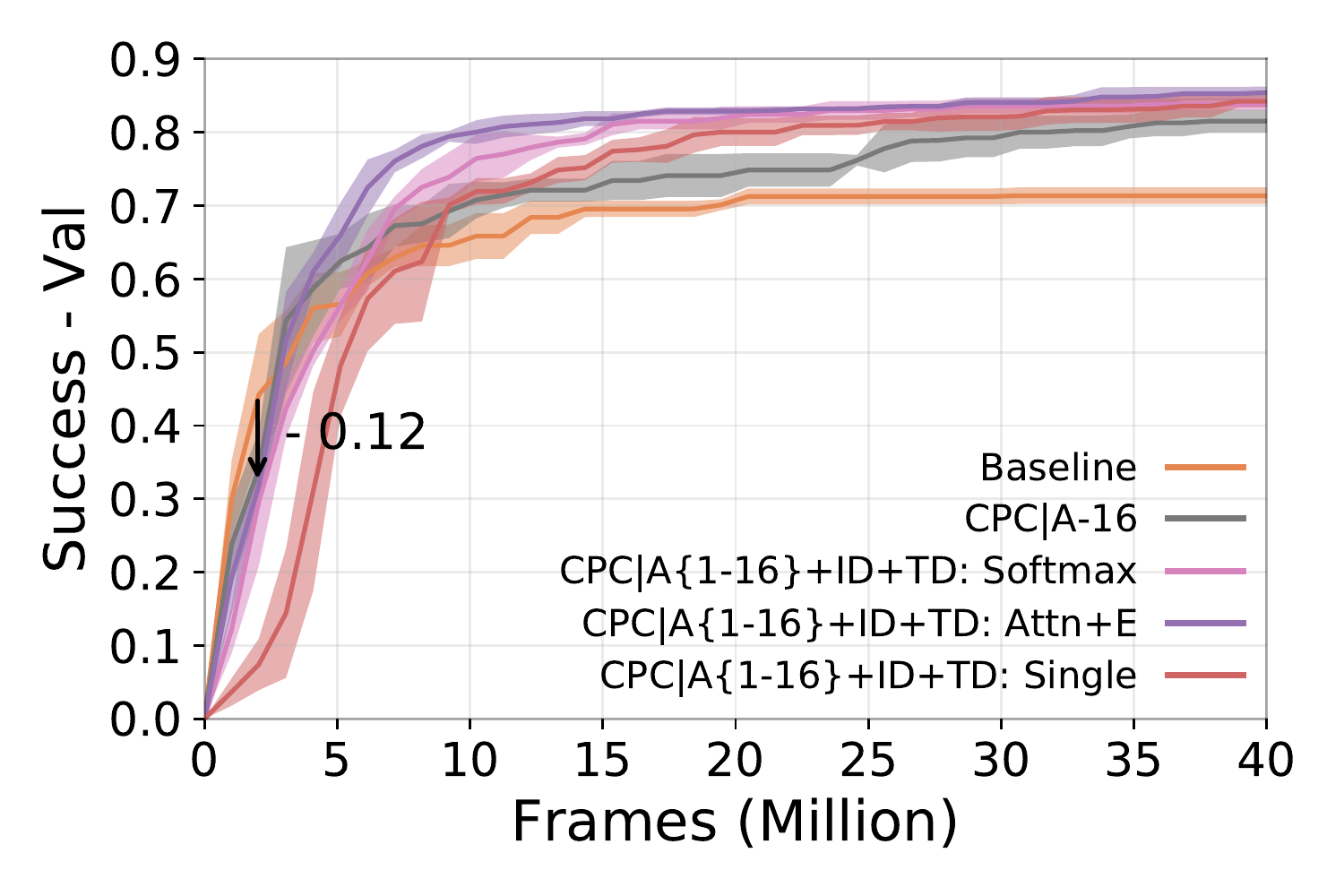}
    \end{subfigure}%
    \begin{subfigure}{.5\textwidth}
      \centering
      \includegraphics[width=\linewidth]{figures/diverse_spl.pdf}
    \end{subfigure}
    \caption{Learning fusion of separate modules in \allcpc+ID+TD: Attn+E recovers the initial ramp-up cost experienced by \allcpc+ID+TD: Single.}
    \label{fig:seven_plots_appendix}
\end{figure}

\newpage
\begin{figure}[t]
    \centering
    \includegraphics[width=12cm]{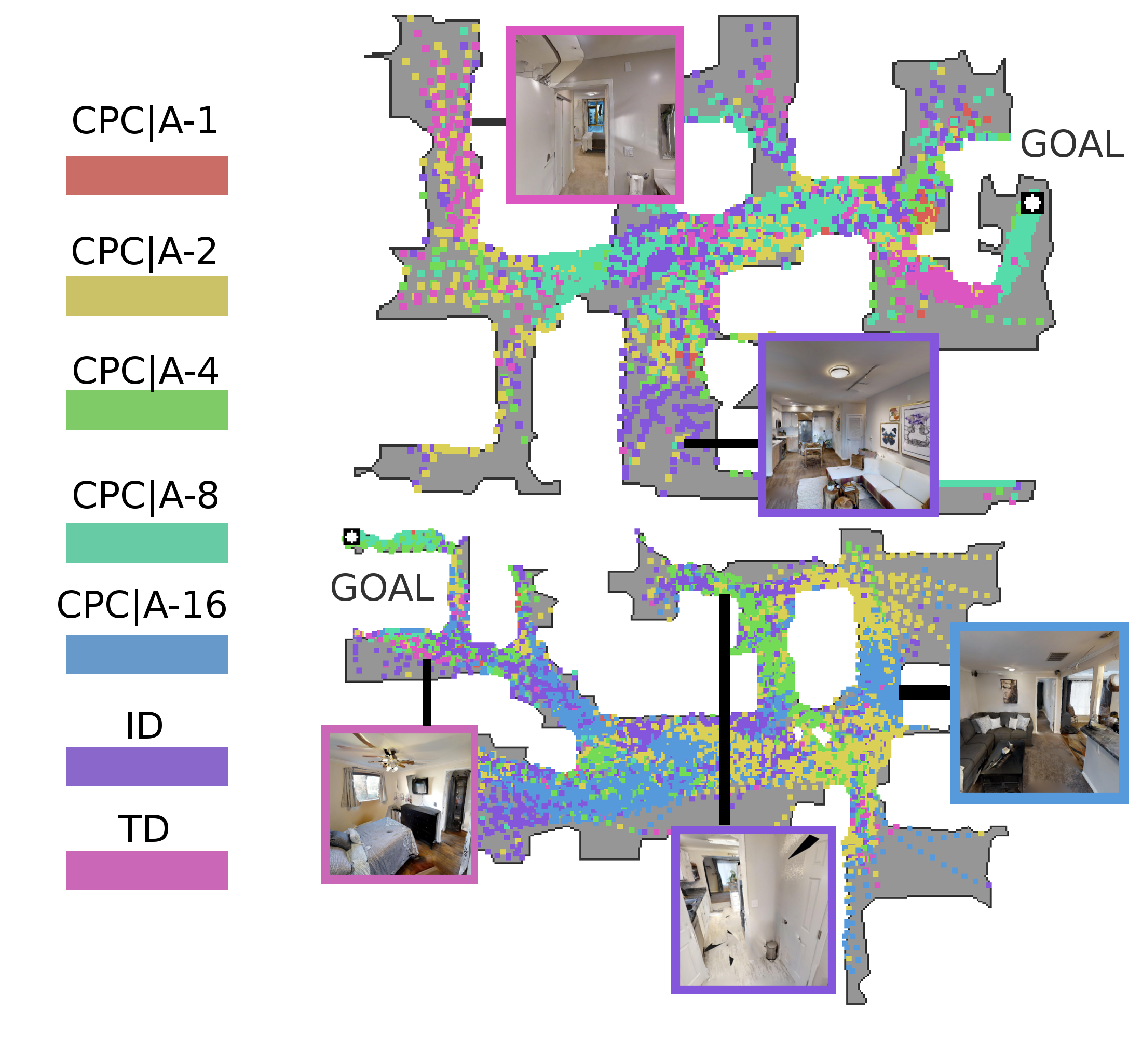}
    \caption{Additional location-based attention weighting visualization. Top: Quantico scene. Bottom: Eastville scene.}
    \label{fig:supp_maps}
\end{figure}
\xhdr{Additional Top Down Map Visualizations}
We also provide top down map visualizations (\figref{fig:supp_maps}) from two more Gibson scenes, Quantico and Eastville. Similar trends prevail as before. The TD task appears to be more activated when beds are in the frame.

\end{document}